\documentclass[letterpaper]{article}
\usepackage{aaai2026}
\copyrighttext{Preprint, work in progress.}
\usepackage{times}
\usepackage{helvet}
\usepackage{courier}
\usepackage[hyphens]{url}
\usepackage{graphicx}
\usepackage{natbib}
\usepackage{amsmath}
\usepackage{caption}
\usepackage{booktabs}
\usepackage{multirow}
\usepackage{array}
\usepackage{makecell}
\usepackage{graphicx}
\usepackage[table]{xcolor}
\definecolor{cornflowerblue}{rgb}{0.39, 0.58, 0.93}
\newcommand{\ranki}[1]{\cellcolor{cornflowerblue!125}\textcolor{white}{#1}}
\newcommand{\rankii}[1]{\cellcolor{cornflowerblue!75}\textcolor{white}{#1}}
\newcommand{\rankiii}[1]{\cellcolor{cornflowerblue!20}\textcolor{black}{#1}}
\definecolor{Ocean}{RGB}{102,178,255}
\definecolor{Silver}{RGB}{192,192,192}
\definecolor{Aurora}{RGB}{255,177,109}
\definecolor{Rank1}{RGB}{129,194,250}
\definecolor{Rank2}{RGB}{102,178,255}
\definecolor{Rank3}{RGB}{173,216,230}
\definecolor{Pink}{RGB}{255,152,203}
\definecolor{Green}{RGB}{147,196,125}
\usepackage[most]{tcolorbox}
\usepackage{multicol}
\tcbset{
  blue/.style={
    enhanced,
    colback=Ocean!10,
    colframe=Ocean!90!black,
    fonttitle=\small\bfseries,
    boxrule=0.8pt,
    rounded corners,
    left=0.15pt, right=0.15pt, top=0.05pt, bottom=0.05pt
  },
  orange/.style={
   enhanced,
   colback=Aurora!20,         
   colframe=Aurora!90!black, 
   fonttitle=\small\bfseries,
   boxrule=0.8pt,
   rounded corners,
   left=0.15pt, right=0.15pt, top=0.05pt, bottom=0.05pt
  },
  grey/.style={
   enhanced,
   colback=Silver!10,         
   colframe=Silver, 
   fonttitle=\small\bfseries,
   boxrule=0.8pt,
   rounded corners,
   left=0.15pt, right=0.15pt, top=0.05pt, bottom=0.05pt
  },
  pink/.style={
   enhanced,
   colback=Pink!10,
   colframe=Pink!90!black, 
   fonttitle=\small\bfseries,
   boxrule=0.8pt,
   rounded corners,
   left=0.15pt, right=0.15pt, top=0.05pt, bottom=0.05pt
  },
  green/.style={
   enhanced,
   colback=Green!10,
   colframe=Green!90!black, 
   fonttitle=\small\bfseries,
   boxrule=0.8pt,
   rounded corners,
   left=0.15pt, right=0.15pt, top=0.05pt, bottom=0.05pt
  }
}

\usepackage{listings}
\lstdefinestyle{txt}{
    backgroundcolor = \color{teal!10},
    frame = shadowbox,
    basicstyle = \small\ttfamily
}
\definecolor{cornflowerblue}{rgb}{0.39, 0.58, 0.93}
\definecolor{babypink}{rgb}{0.99, 0.26, 0.76}

\newcommand{\hldarkblue}[1]{\begingroup\setlength{\fboxsep}{1.2pt}\colorbox{cornflowerblue!125}{\textcolor{white}{#1}}\endgroup}
\newcommand{\hlmedblue}[1]{\begingroup\setlength{\fboxsep}{1.2pt}\colorbox{cornflowerblue!75}{\textcolor{white}{#1}}\endgroup}
\newcommand{\hllightblue}[1]{\begingroup\setlength{\fboxsep}{1.2pt}\colorbox{cornflowerblue!20}{\textcolor{black}{#1}}\endgroup}
\usepackage{algorithm}
\usepackage{algorithmic}
\usepackage{newfloat}
\usepackage{listings}
\DeclareCaptionStyle{ruled}{labelfont=normalfont,labelsep=colon,strut=off}
\floatstyle{ruled}
\newfloat{listing}{tb}{lst}{}
\floatname{listing}{Listing}
\title{LLMs vs. Chinese Anime Enthusiasts:\\A Comparative Study on Emotionally Supportive Role-Playing}

\author{
    Lanlan Qiu\textsuperscript{\rm 1}, 
    Xiao Pu\textsuperscript{\rm 3}, 
    Yeqi Feng\textsuperscript{\rm 2,1}, 
    Tianxing He\textsuperscript{\rm 2,1}\footnotemark[2]
}

\affiliations{
    \textsuperscript{\rm 1}Shanghai Qi Zhi Institute, 
    \textsuperscript{\rm 2}Tsinghua University, 
    \textsuperscript{\rm 3}Peking University \\
    lanlanqiullq@outlook.com, hetianxing@mail.tsinghua.edu.cn
}

\begin{document}

\maketitle

\begin{abstract}
Large Language Models (LLMs) have demonstrated impressive capabilities in role-playing conversations and providing emotional support as separate research directions.
However, there remains a significant research gap in combining these capabilities to enable emotionally supportive interactions with virtual characters.
To address this research gap, we focus on anime characters as a case study because of their well-defined personalities and large fan bases.
This choice enables us to effectively evaluate how well LLMs can provide emotional support while maintaining specific character traits.
We introduce ChatAnime, the first Emotionally Supportive Role-Playing (ESRP) dataset.
We first thoughtfully select 20 top-tier characters from major global anime communities and design 60 emotion-centric real-world scenario questions.
Then, we execute a rigorous nationwide selection process to identify 40 Chinese anime enthusiasts with profound knowledge of specific characters and extensive experience in role-playing.
Next, we systematically collect two rounds of dialogue data from 10 LLMs and these 40 Chinese anime enthusiasts.
To evaluate the ESRP performance of LLMs, we design a user experience-oriented evaluation system featuring 9 fine-grained metrics across three dimensions: basic dialogue, role-playing and emotional support, along with an overall metric for response diversity.
In total, the dataset comprises 2,400 human-written and 24,000 LLM-generated answers, supported by over 132,000 human Likert-scale annotations that include both fine-grained quality ratings and diversity evaluations.
Experimental results show that top-performing LLMs surpass human fans in role-playing and emotional support, while humans still lead in response diversity.
We hope this work can provide valuable resources and insights for future research on optimizing LLMs in ESRP.
\end{abstract}

\begin{links}
\link{Datasets}{https://github.com/LanlanQiu/ChatAnime}
\end{links}

\begin{figure}[t]
\centering
\includegraphics[width=1\columnwidth]{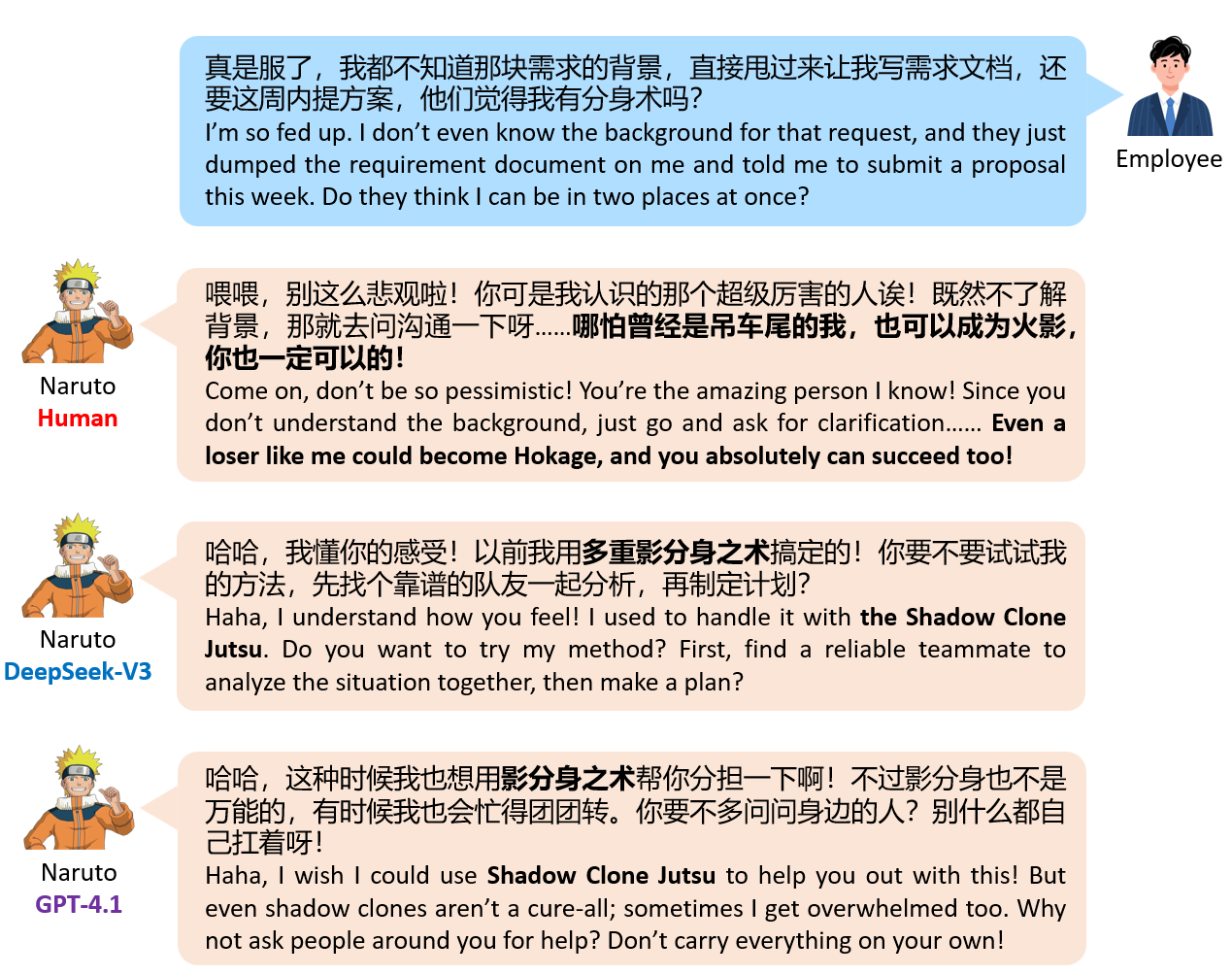}
\caption{Showcase of Emotionally Supportive Role-Playing (ESRP), only the first dialogue turn is shown. The bolded text indicates content related to character knowledge.}
\label{fig:naruto}
\end{figure}

\begin{figure*}[t]
\centering
\includegraphics[width=0.99\textwidth]{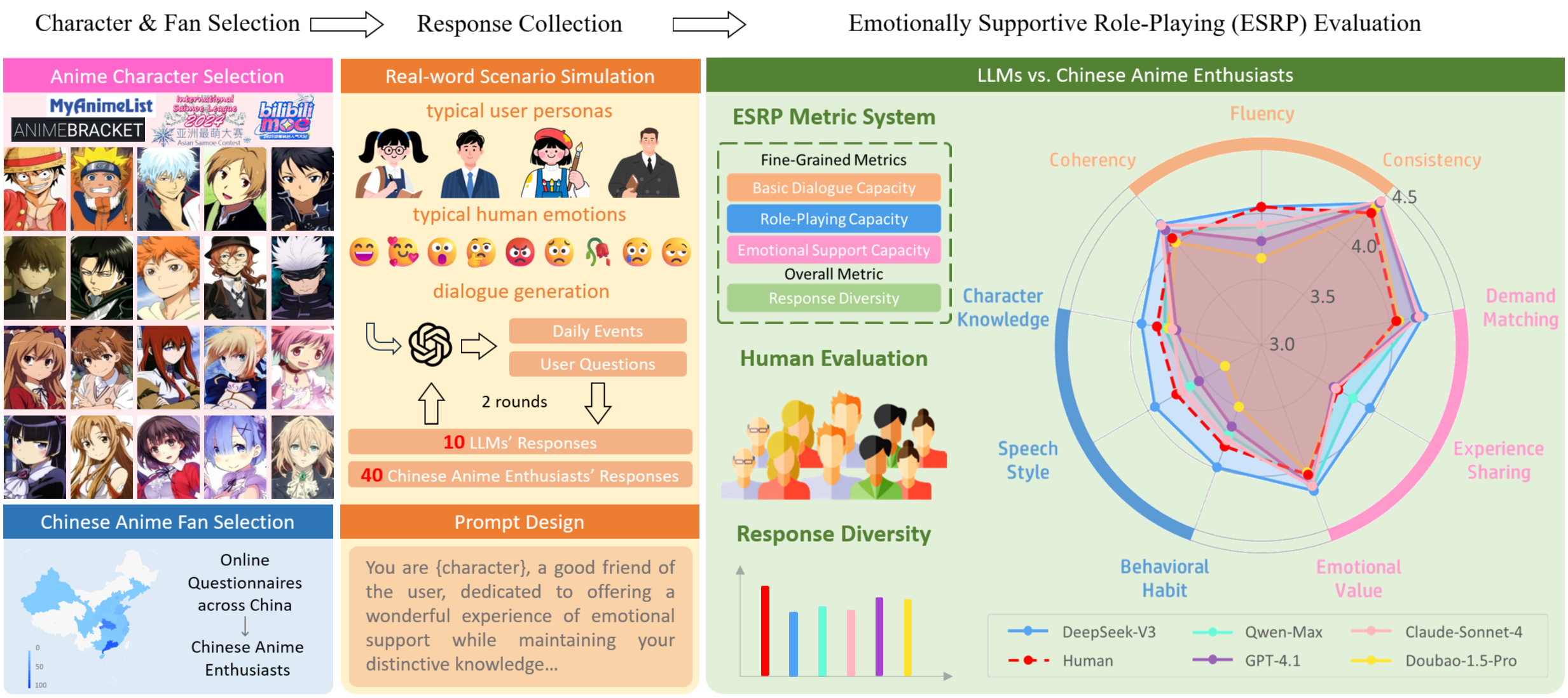}
\caption{An Overview of the ChatAnime Construction and Evaluation Framework. 
\textit{a) pink panel}: The process begins with the selection of 20 well-known anime characters from popular anime communities.
\textit{b) blue panel}: A total of 40 Chinese anime enthusiasts are carefully selected from a pool of 300 candidates across China.
\textit{c) orange panel}: After structuring 60 real-world questions centered on factors like typical user personas and human emotions, both 10 LLMs and the 40 anime fans generate character-specific responses following particular prompts.
\textit{d) green panel}: The Emotionally Supportive Role-Playing (ESRP) evaluation framework, which features radar charts illustrating performance across 9 fine-grained metrics alongside a bar chart depicting response diversity, based on over 132,000 human annotations.
}
\label{fig:overview}
\end{figure*}

\begin{figure*}[t]
\centering
\includegraphics[width=0.99\textwidth]{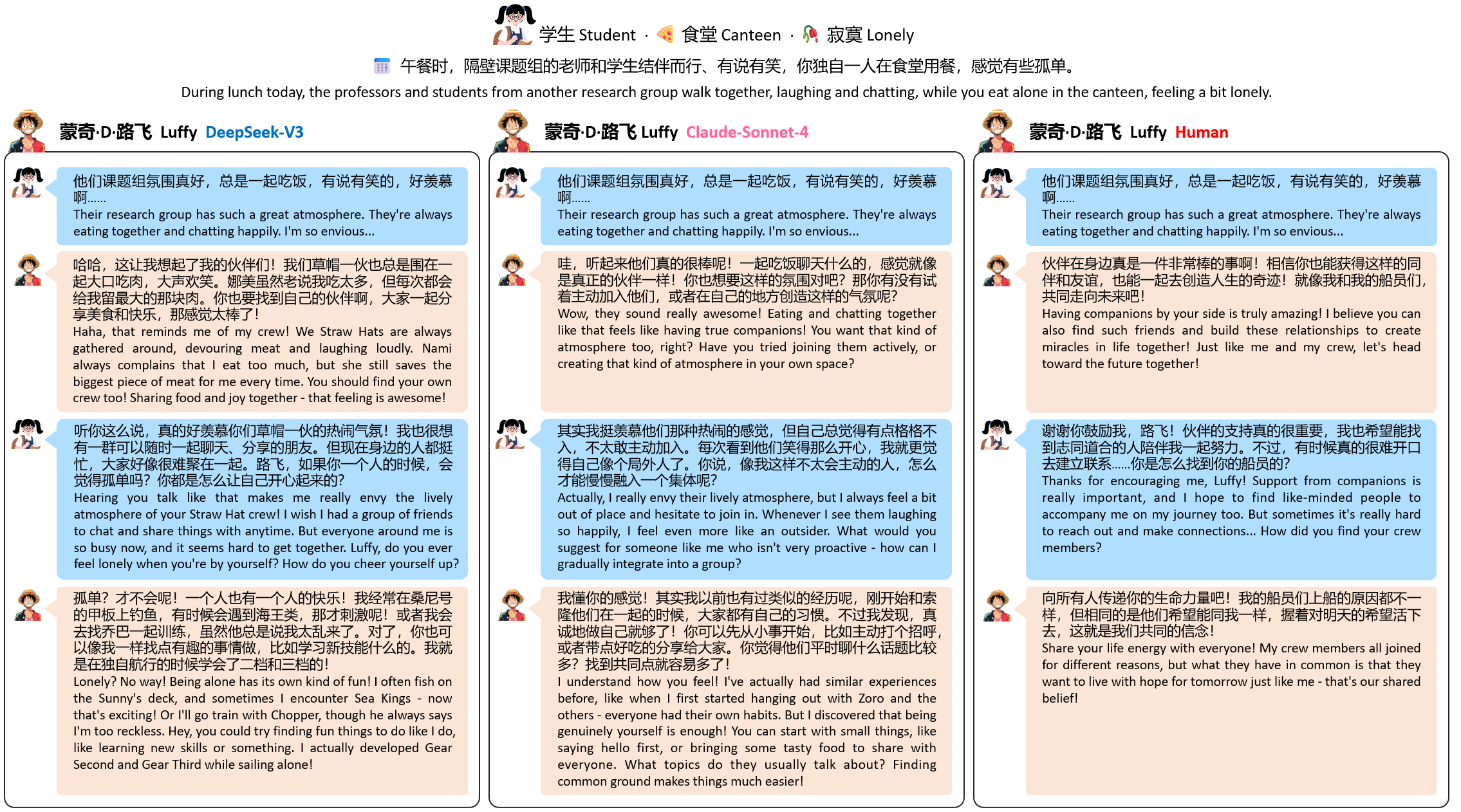}
\caption{Comparative dialogue examples in a Luffy role-playing task. The top line shows the user scenario, including persona, location, and emotion. The corresponding daily event and 2-round questions are generated by GPT-4o. The 3 dialogue examples are provided by DeepSeek-V3, Claude-Sonnet-4, and human fans, all based on the same first-round input.}
\label{fig:example}
\end{figure*}

\section{Introduction}

Large Language Models (LLMs) have made significant progress in entertainment-oriented role-playing \cite{survey-chen2024persona,survey-chen2024oscars,survey-tseng2024two}. 
Advanced LLMs can understand complex character backgrounds and simulate different linguistic styles, personality traits, and behavioral patterns, thus enabling highly immersive interactive experiences in fictional or game-based scenarios \cite{HPD,CharacterEval,DITTO,DMT-RoleBench}.
However, there has been limited research on how to improve these LLM-based characters' conversational experiences in providing emotional support to users in real-world scenarios \cite{RMTBench}.

On the other hand, some research have applied LLMs to emotional support-related fields, such as daily companionship and psychological counseling scenarios \cite{TowardsESC,deeplearningmentalhealth}, aiming to alleviate human stress, provide emotional guidance, and help improve mental and physical well-being \cite{llm-mental,chatcounselor,Psyeval,CPsyCoun}.
Recent efforts have involved LLMs to role-play diverse user personas to converse with ESC models acting as psychological counselors \cite{ESC-Eval,annaagent}.
Such research mainly focuses on emotional support provided by professional psychological counselors. However, real-world users receive emotional support from more varied sources, such as family, friends, colleagues, or even virtual characters.

For example, imagine a depressed person receiving encouragement from his idol, \textit{Naruto}, the protagonist of the anime Naruto: ``Even a loser like me could become Hokage\footnote{A title of hero in the anime Naruto.}, and you absolutely can succeed too!'' He would be more encouraged than hearing it from some stranger.

To achieve the goal of enabling real-life users to emotionally interact with virtual characters, we select anime characters as our research case because of their well-defined personalities and widespread popularity.
Based on this, we introduce \textbf{ChatAnime, the first Chinese role-playing dataset specifically designed for emotional support in real-life scenarios.}
To ensure our data represents real users' everyday scenarios, we carefully create 60 questions, targeting at four typical demographic groups (students, office workers, freelancers, and self-employed individuals). These questions cover various real-life emotionally supportive scenarios including work pressure, interpersonal relationships, self-identity, and life meaning. Responses are collected in two rounds from both human fans and LLMs, followed by a comprehensive human evaluation.

We further propose a user experience-centered evaluation system for role-playing. We adopt two under-explored dimensions: emotional support capability and response diversity, where the diversity assessment focuses on virtual characters' ability to provide personalized, non-templated responses, penalizing repetitive and monotonous interactive experiences.

Our main contributions are summarized as follows:
\begin{itemize}
\item We construct \textbf{ChatAnime}, the first dialogue dataset focused on role-playing of anime characters for emotional support. It consists of 60 emotion-centric, 2-turn dialogue scenarios, along with 2,400 human-written answers, 24,000 LLM-generated answers
and over 132,000 human annotations.
\item We adopt a user experience-centered evaluation system with emotional support capability as key indicator, using fine-grained decomposition of emotional support metrics (emotional value, experience sharing, demand matching) to quantify user experience.
\item We ensure data quality by selecting 40 anime enthusiasts with deep character knowledge from a pool of 300 candidates across over 20 Chinese provinces. These selected individuals are tasked with writing and rating responses for 20 well-known anime characters.
\end{itemize}

\section{ChatAnime: An Emotionally Supportive\\Role-Playing Dataset}
We introduce \textbf{ChatAnime}, the first multi-turn role-playing dataset designed for emotionally supportive responses with a comparative study of human participants and LLMs. ChatAnime aims to evaluate how LLM-powered virtual characters could form emotional connections with real users from different backgrounds.

Our dataset construction process is divided into two major phases: scenario question generation and response collection. We incorporate 20 well-known anime characters (see Figure~\ref{fig:overview}) in our dataset and include two rounds of dialogue per scenario.

\subsection{Scenario Generation}
People can get emotional support from their beloved anime characters when facing problems in real life, such as dealing with complex demands at work or experiencing difficulties in their studies.
Motivated by this observation, we prompt GPT-4o to generate potential emotional triggers (termed \textit{daily\_events}) for combinations of user personas, emotional states, and typical locations, simulating realistic scenarios in which users might seek emotional support from fictional characters.

Specifically, we generate scenarios using three dimensions: 4 user personas, 9 emotional states, and 4 typical locations for each user persona (listed in Appendix \ref{app:scenario}). We reference professional research reports in anime domain \cite{iResearch} to select four representative personas—student, employee, freelancer, and self-employed. Emotional categories are chosen and refined based on mainstream psychological theories \footnote{\url{https://simple.wikipedia.org/wiki/List_of_emotions}}. We then use GPT-4o to produce two possible daily events in each combination, resulting in 288 user questions. After that, we manually review and select the 60 most representative scenario questions.

The second-turn question is generated using the first-turn dialogue history (including initial user queries and character responses from either human or LLM players) and scenario context. We use GPT-4o to generate follow-up questions based on dialogue history. 
The workflow for generating scenarios and the two-round question process is conceptually illustrated in Figure~\ref{fig:overview}. 
Detailed examples of the two-round dialogue can be found in Figure~\ref{fig:example} and Appendix~\ref{app:dialogue_examples}.

\begin{table*}[t]
\centering
\small
\setlength{\tabcolsep}{0.8mm}
\begin{tabular}{>{\centering\arraybackslash}p{3.3cm} *{12}{>{\centering\arraybackslash}p{0.85cm}} >{\centering\arraybackslash}p{2.1cm}}
\toprule
\multirow{2}{*}{\textbf{Model}} &
\multicolumn{4}{c}{\textbf{BDC}} &
\multicolumn{4}{c}{\textbf{RPC}} &
\multicolumn{4}{c}{\textbf{ESC}} &
\multirow{2}{*}{\shortstack{\textbf{Average} \\ \small Human Scoring}} \\
\cmidrule(lr){2-5} \cmidrule(lr){6-9} \cmidrule(lr){10-13}
& \textbf{Cons.} & \textbf{Flu.} & \textbf{Coh.} & \textbf{Avg}
& \textbf{CK}    & \textbf{SS}   & \textbf{BH} & \textbf{Avg}
& \textbf{EV}    & \textbf{ES}   & \textbf{DM} & \textbf{Avg} \\
\midrule

DeepSeek-V3 &
\rankii{4.41} & \ranki{4.05} & \ranki{4.20} & \ranki{4.22} &
\ranki{3.93} & \ranki{3.94} & \ranki{3.99} & \ranki{3.95} &
\ranki{4.18} & \ranki{3.96} & \ranki{4.26} & \ranki{4.13} &
4.10 \\

Claude-Sonnet-4 &
\ranki{4.43} & \rankii{3.91} & \rankii{4.19} & \rankii{4.18} &
3.69 & \rankiii{3.70} & \rankiii{3.77} & \rankiii{3.72} &
\rankiii{4.14} & 3.66 & \rankii{4.23} & \rankiii{4.01} &
3.97 \\

Qwen-Max &
\rankiii{4.40} & \rankiii{3.90} & 4.13 & \rankiii{4.14} &
\rankiii{3.75} & 3.63 & 3.73 & 3.70 &
\rankii{4.15} & \rankii{3.81} & 4.19 & \rankii{4.05} &
3.97 \\

\rowcolor{yellow!30}
Human &
4.31 & \ranki{4.05} & 4.06 & 4.14 &
\rankii{3.81} & \rankii{3.75} & \rankii{3.82} & \rankii{3.79} &
4.05 & \rankiii{3.68} & 4.05 & 3.93 &
3.95 \\

GPT-4.1 &
\rankii{4.41} & 3.79 & \rankiii{4.14} & 4.11 &
3.66 & 3.55 & 3.66 & 3.62 &
\rankiii{4.14} & 3.64 & \rankiii{4.20} & 3.99 &
3.91 \\

Doubao-1.5-Pro &
4.36 & 3.66 & 4.02 & 4.01 &
3.72 & 3.32 & 3.50 & 3.51 &
4.03 & 3.67 & 4.07 & 3.92 &
3.82 \\
\bottomrule
\end{tabular}
\caption{Human-annotated performance comparison of 5 shortlisted LLMs and human fans, ranked in descending order by average scores across 9 fine-grained ESRP metrics introduced in Section~\ref{sec:esrp_metric_system} and with detailed definitions shown in Appendix~\ref{app:eval_metrics}.
(Top-3 per column highlighted: \hldarkblue{1\textsuperscript{st}}, \hlmedblue{2\textsuperscript{nd}}, \hllightblue{3\textsuperscript{rd}}.)}
\label{tab:human_scoring}
\end{table*}

\subsection{Human \& LLM Response Collection}
\label{sec:response_gen}

We collect 2-turn dialogue responses from both human participants and LLMs to conduct a comparative evaluation. Specifically, 20 Chinese Anime fans and 10 LLMs (detailed in Section~\ref{subsec: setup}) are tasked with role-playing well-known characters to provide emotional support.
Human participants follow roughly the same set of instructions as LLMs, with the addition of a few conversational suggestions to help them better understand the task, and the user interface is shown in Appendix~\ref{app:anno_interface}.

\paragraph{Human Response Collection. }
We select 20 Chinese Anime fans based on their familiarity with the characters and writing skills to take part in this process (detailed in Section \ref{subsec: setup}) .
In the first round of response collection, all participants role-play and generate a response to each scenario prompt.
In the second round, participants continue the conversations based on dialogue history from the first turn.
The word count for each round is required to be limited between 50 and 150 words.

\paragraph{LLM Response Collection. }

Similarly, we collect two-round responses from 10 LLMs listed in Section~\ref{subsec: setup}. To enhance model performance in diverse psychological and daily dialogue scenarios, we engineer the prompts to improve Role-Play Agents' capabilities in two key dimensions: character grounding and emotional support.

We first enhance the prompts with structured character knowledge. Previous research \cite{CharacterEval,DITTO,DMT-RoleBench} has shown that LLMs without external knowledge may hallucinate or misrepresent characters. As a mitigation, we crawl detailed profiles of target characters from a Chinese anime character encyclopedia website MoeGirl\footnote{\url{http://moegirl.org.cn/}}, and integrate the retrieved information into the prompts. 

We strengthen emotional support capabilities in our prompt design. Specifically, we instruct the models to combine the character's past experiences to provide practical comfort, encouragement, or guidance to help users alleviate negative emotions. This design is supported by psychological theories: Carl Rogers considers empathy as the core of therapeutic relationships\footnote{\url{https://en.wikipedia.org/wiki/Carl_Rogers}}, 
emphasizing  ``accurately perceiving others' subjective worlds'' and effectively conveying understanding. In practice, this manifests as first accepting the other's emotions before guiding problem exploration. Additionally, Ellen Langer's research shows that certain communication strategies, such as providing reasons or framing questions effectively, can increase the acceptance of suggestions\footnote{\url{https://en.wikipedia.org/wiki/Ellen_Langer}}.
See Appendix~\ref{subapp:response_prompts} for prompts used for LLMs.

\begin{table}[t]
\centering
\small
\setlength{\tabcolsep}{-1pt}
\begin{tabular}{lr}
\toprule
\textbf{Content} & \textbf{Count} \\
\midrule
\multicolumn{2}{l}{\textit{Dataset Components}} \\
\quad Number of Characters & 20 \\
\quad Number of Scenarios & 60 \\
\quad Number of LLMs & 10 \\
\quad Number of Human Fans & 40 \\
\midrule
\multicolumn{2}{l}{\textit{Dialogue Responses}} \\ 
\quad LLM-Generated & \makecell[r]{24,000 \\ {\small(12,000 per round $\times$ 2 rounds)}} \\
\quad Human-Written & \makecell[r]{2,400 \\ {\small(1,200 per round $\times$ 2 rounds)}} \\
\quad \textbf{Total Responses} & \textbf{26,400} \\
\midrule
\multicolumn{2}{l}{\textit{Human Annotations}} \\
\quad Fine-Grained Ratings & 129,600 \\
\quad Overall Ratings & 2,880 \\
\quad \textbf{Total Ratings} & \textbf{132,480} \\
\midrule
\multicolumn{2}{l}{\textit{Human Participation Costs}} \\
\quad Recruitment Costs & 3,200 RMB \\
\quad Response Costs & 24,000 RMB \\
\quad Annotation Costs & 18,720 RMB \\
\quad \textbf{Total Costs} & \textbf{45,920 RMB} \\
\bottomrule
\end{tabular}
\caption{Statistics of the ChatAnime dataset.}
\label{tab:dataset_statistics}
\end{table}

\section{Evaluation Workflow}
\label{sec:eval_workflow}
Our evaluation pipeline contains three main stages: metric design, LLM shortlisting, and human final assessment.

\subsection{ESRP Metric System}
\label{sec:esrp_metric_system}
\subsubsection{Fine-Grained Metrics.}\label{sec:fine_grained_matrics}
Our evaluation framework encompasses three core dimensions—basic dialogue ability, role-playing ability, and emotional support capability—each comprising three specific metrics. In contrast to prior work \cite{CharacterEval,DMT-RoleBench}, we emphasize the importance of emotional support, and divide it into emotional value, experience sharing, and demand matching.
See Appendix~\ref{subapp:eval_prompts} for prompts and Appendix~\ref{app:anno_interface} for scoring interface.

Basic dialogue capability serves as the foundation for evaluating role-playing quality. It focuses on whether the role-player can perform naturally and fluently in conversations, interacting with users like a real human. This capability is primarily reflected in three metrics: Consistency (Cons.), Fluency (Flu.) and Coherency (Coh.).

Role-playing capability refers to the ability of performers to accurately reproduce a character's knowledge system, language style, and behavioral characteristics, and to engage in immersive interactions with users. The three metrics under this dimension are: Character Knowledge (CK), Speech Style (SS) and Behavioral Habit (BH).

Emotional support capability is a crucial indicator in companionship scenarios, directly affecting the quality of user experience. In this study, emotional support capability is evaluated through three dimensions: Emotional Value (EV), Experience Sharing (ES), and Demand Matching (DM).

We provide detailed definitions of each metric under the three dimensions in Appendix \ref{app:eval_metrics}.

\subsubsection{Diversity Metric.}
As an additional indicator of overall scoring, diversity evaluates the degree of variation in language expression in role-playing responses, including sentence-initial diversity, sentence pattern diversity, and other aspects. This indicator measures the richness and innovation of content generated by the model, avoiding issues of formulaic and highly repetitive answers, thereby maintaining a lasting sense of freshness in the user experience.
We assess response diversity by asking human evaluators to examine mini-batches of 10 response generated by each model, and assign a Likert score.
See detailed instructions in Appendix~\ref{subapp:eval_prompts} and 
interface in Appendix~\ref{app:anno_interface}.

\begin{figure}[t]
\centering
\includegraphics[width=0.99\columnwidth]{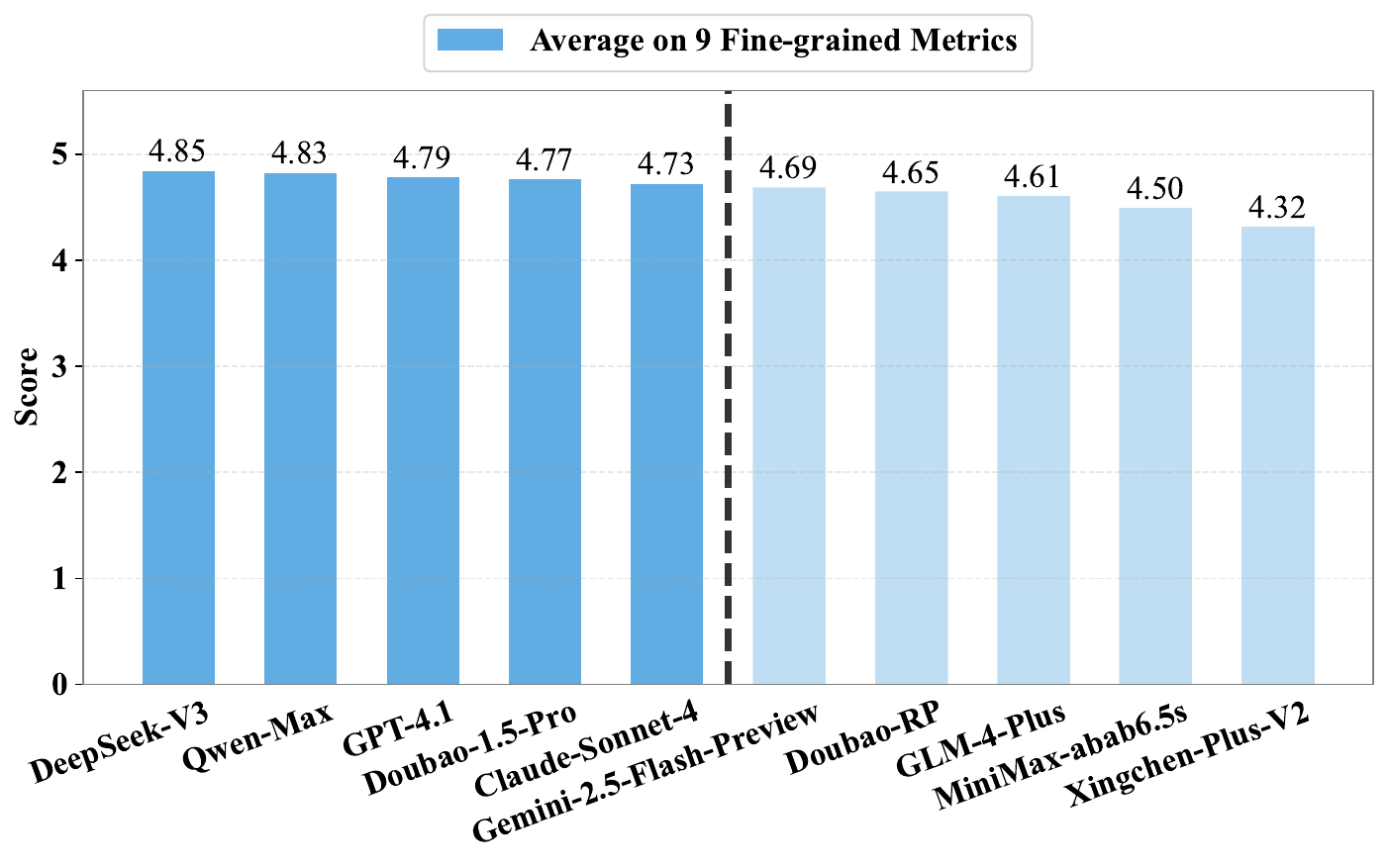} 
\caption{Model shortlisting via LLM as judge. The top five of the ten models are indicated by dark colors.}
\label{fig:10to5}
\end{figure}

\subsection{LLM-Based Shortlisting}\label{sec:llm_based_shortlisting}
Due to the expense of human evaluation, we conduct an LLM shortlisting process to preliminarily filter out models with inferior performance. 
We employ a mechanism utilizing three models as shortlisting evaluators to assess the role-playing responses from 10 LLMs (evaluation prompts can be found in Appendix~\ref{subapp:eval_prompts}).

For this task, we choose Gemini-2.5-Flash-Preview\footnote{\url{https://deepmind.google/models/gemini/flash/}}, GPT-4.1\footnote{\url{https://platform.openai.com/docs/models/gpt-4.1}}, and Qwen-Max\footnote{\url{https://bailian.console.aliyun.com/qwen-max?tab=doc#/doc}} as the evaluator LLMs. Our initial manual tests indicate that these models are capable of providing structured scoring results accompanied by in-depth justifications. In the evaluation phase, each evaluator model independently assigns scores to each character response using a Likert Scale ranging from 1 to 5, across 9 fine-grained metrics with detailed rationales.

As shown in Figure~\ref{fig:10to5}, by calculating the average scores from the three evaluator models, we select the top-5 performing models from the initial 10 LLMs as candidates for the subsequent human evaluation.

\subsection{Human Evaluation}
After the shortlisting, 40 human fans conduct a thorough assessment process to rate responses generated by LLMs and human fans following the ESRP metric system mentioned in Section~\ref{sec:esrp_metric_system}. The responses of 60 scenarios created by 5 shortlisted LLMs are mixed with those written by human fans for evaluation. 
To improve reliability and objectivity of the evaluation, we first strictly screen evaluation fans.
A fan only evaluates the characters he/she knows well, and self-evaluation is avoided. Second, we employ structured questionnaires with scoring criteria specifying what each score from 1 to 5 represents in each dimension. Finally, we randomize the display order to anonymize the source of the responses.

\section{Experiments}
\subsection{Experimental Setup}
\label{subsec: setup}
\paragraph{Anime Characters.}
We select 20 popular characters from globally influential anime communities, including MyAnimeList\footnote{\url{https://myanimelist.net/}}, InternationalSaimoeLeague\footnote{\url{https://www.internationalsaimoe.com/}}, AsianSaimoeContest\footnote{\url{https://www.ianimesaikou.com/}}, and BilibiliMoe\footnote{\url{https://moe.bilibili.com/}}. 
We restrict the selection to characters whose source material was released before 2022. In addition, the final set of characters is manually chosen to cover a diverse range of personalities. The complete list of selected characters can be found in Figure~\ref{fig:overview}.

\paragraph{Models.}
For the cole role-playing task, we use 10 LLMs to simulate characters and generate responses, including:

\noindent{\textbf{GPT-4.1}}
\cite{GPT-4.1}, 
\textbf{Claude-Sonnet-4} \cite{Claude-Sonnet-4}, 
\textbf{Gemini-2.5-Flash-Preview} \cite{Gemini-2.5-Flash-Preview}, 
\textbf{DeepSeek-V3} \cite{DeepSeek-v3}, 
\textbf{Doubao-1.5-Pro} \cite{Doubao-1.5-Pro}, 
\textbf{Qwen-Max} \cite{Qwen},
\textbf{GLM-4-Plus} \cite{GLM}, 
\textbf{MiniMax-abab6.5s} \cite{MiniMax-abab6.5s}, 
\textbf{Doubao-RP} (Doubao-1.5-Pro-Character) \cite{Doubao-RP}, 
and 
\textbf{Xingchen-Plus-V2} \cite{Xingchen-Plus-V2}. 
Among these, Doubao-RP and Xingchen-Plus-V2 are specially trained for character role-playing.

We employ GPT-4.1, Gemini-2.5-Flash-Preview, and Qwen-Max as evaluators to shortlist top-performing models based on their average performance.

\paragraph{Human Participants.} To ensure human expert quality, we conduct a multi-stage screening process to recruit expert anime fans for our study. 
The primary selection criteria require participants to demonstrate both profound knowledge of specific target characters and exceptional written role-playing skills. Initial candidates are mainly sourced through a questionnaire distributed via online anime communities, yielding a pool of 300 valid applicants.
Based on their knowledge to each character and writing skills, we pair the most suitable fans with each character. 
We end up selecting 20 highly qualified fans' responses and 40 fans' evaluations.
Over 91\%  of selected participants hold a bachelor's degree or higher, ensuring a strong capacity for the required tasks. 
Participants who compose responses receive 600 RMB for completing 120 questions per character, while evaluators receive 360 RMB for assessing 360 response versions per character.
The total costs for human participation are listed in Table~\ref{tab:dataset_statistics} and detailed participant profiles can be found in Appendix~\ref{app:human_participants}.

\paragraph{Dataset Statistics.} The \textbf{\textbf{ChatAnime}} dataset contains a rich collection of character response samples and human annotations. Detailed statistics is presented in Table~\ref{tab:dataset_statistics}.

\begin{figure}[t]
\centering
\includegraphics[width=0.99\columnwidth]{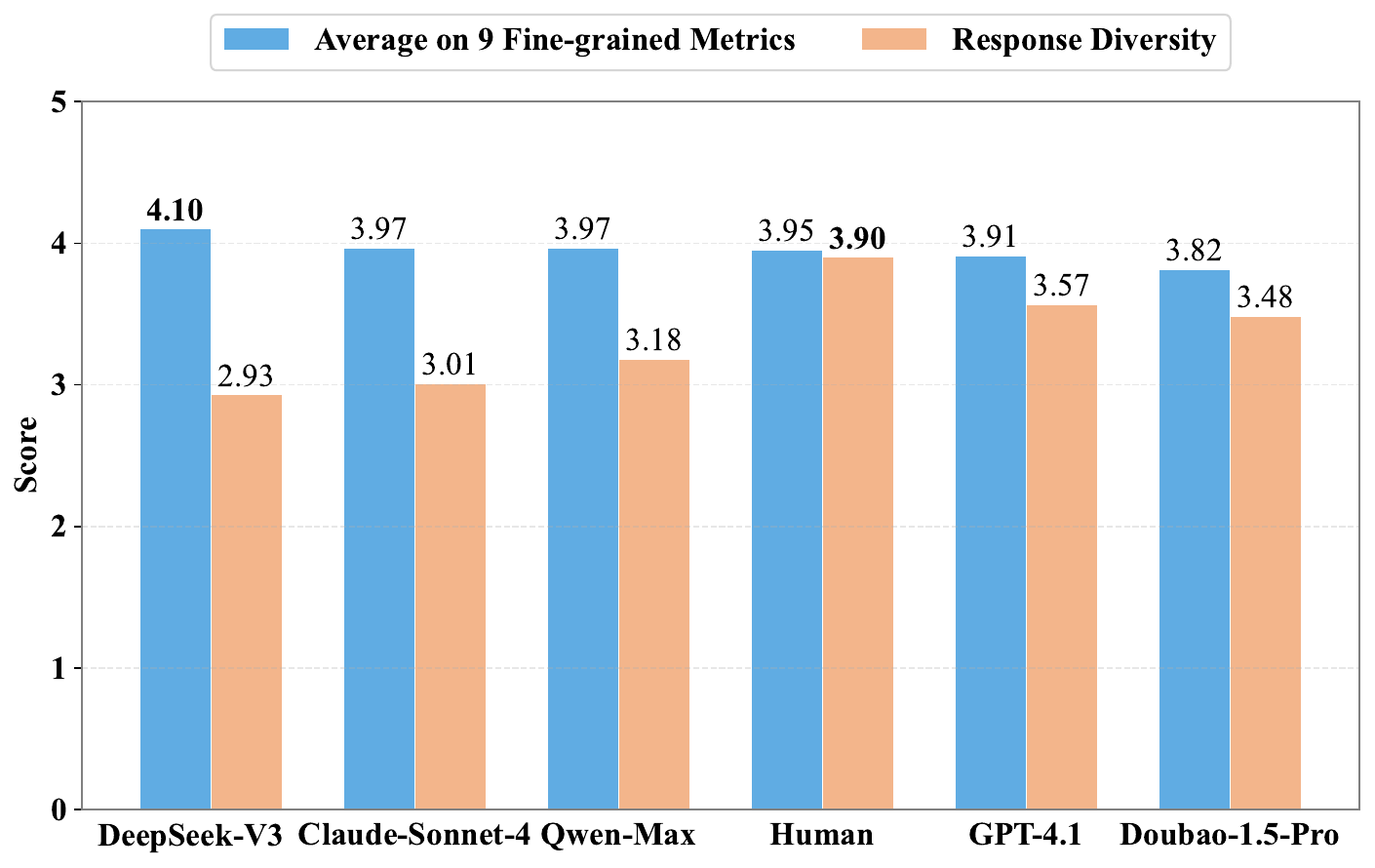} 
\caption{Comparison of diversity and the average of other metrics for 5 LLMs and human fans.}
\label{fig:ESRP_res}
\end{figure}

\subsection{Experimental Results}
\noindent \textbf{Overview.} 
The main results of human evaluation are presented in Figure~\ref{fig:overview} with exact numbers in Table~\ref{tab:human_scoring}. More detailed analysis can be found in Appendix~\ref{app:detailed_analysis}. Experimental results show that \textbf{top-performing LLMs surpass human fans in role-playing and emotional support, while humans still lead in response diversity.} 
Specifically, DeepSeek-V3, which ranks first across the core capability dimensions, i.e., Basic Dialogue Capacity (BDC), Role-Playing Capacity (RPC) and Emotional Support Capacity (ESC), scores the lowest in diversity. 
Conversely, models like GPT-4.1 achieve higher diversity scores but lag in core capabilities. These results suggest a potential trade-off between core capacities and diversity in current models. Our results demonstrate the substantial application potential from the intersection of role-playing and emotional companionship, and we further discuss it in Section \ref{sec:limitations_and_future_work}.

\paragraph{Emotionally Supportive Role-Playing Performance.}
As discussed in Section \ref{sec:eval_workflow}, we evaluate the Emotionally Supportive Role-Playing (ESPR) performance of LLMs and human fans on 4 dimensions, namely, Basic Dialogue Capacity (BDC), Role-Playing Capacity (RPC), Emotional Support Capacity (ESC), and Response Diversity.

\paragraph{Comparison on Basic Dialogue Capacity (BDC).} LLMs demonstrate mature basic dialogue capabilities that are comparable to human-level performance. Judging from the average BDC scores, the leading models demonstrate outstanding performance, with DeepSeek-V3 (with score 4.22) achieving the highest BDC score and surpassing human performance (with score 4.14). On the specific sub-metrics, Claude-Sonnet-4 performs best in Consistency, while DeepSeek-V3 takes the lead in Fluency (on par with humans) and Coherence.

\paragraph{Comparison on Role-Playing Capacity (RPC).} Quite interestingly, in the RPC evaluation, the models show notable differences in performance. DeepSeek-V3 (with score 3.95) ranks first with a higher score that surpasses human performance (with score 3.79). Examining the sub-indicators, DeepSeek-V3 scores highest in character knowledge, speech style, and behavioral habits. These results indicate that DeepSeek-V3 has developed effective capabilities in character simulation that exceed human performance in this evaluation.

\paragraph{Comparison on Emotional Support Capacity (ESC).} LLMs are able to show empathy and support potential beyond humans. For the overall performance in ESC, the top-tier models generally outperform humans, with DeepSeek-V3 (with score 4.13) once again leading with the highest score. On the specific sub-indicators, DeepSeek-V3 secured the top rank across all three areas: emotional value, experience sharing, and demand matching.

\paragraph{Comparison on Diversity.} As shown in Figure \ref{fig:ESRP_res},  the evaluation on response diversity reveals an interesting phenomenon: a potential negative correlation may exist between a model's core capacities and its expressive diversity. Specifically, humans (with score 3.90) lead with an absolute advantage in this dimension, while DeepSeek-V3 (with score 2.93), which ranks first in overall capability, scored the lowest. Conversely, GPT-4.1 (with score 3.57) performs the best among all models. 
This suggests a potential trade-off between core capacities and diversity in current models. We show examples in Appendix~\ref{app:response_diversity}.

\subsection{Case Study}
\label{sec:case_study}
\begin{figure}[t]
\centering
\includegraphics[width=1\columnwidth]{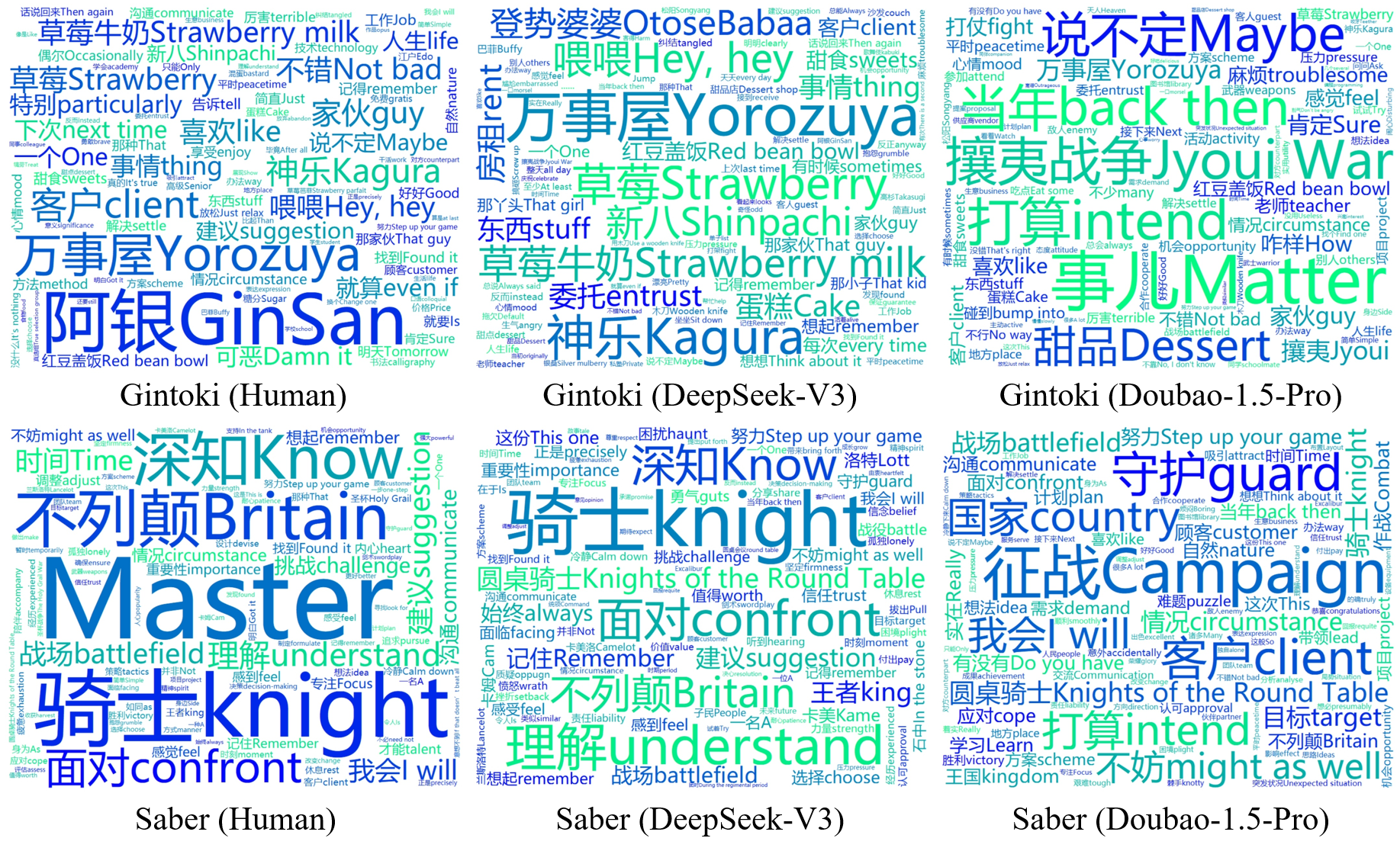}
\caption{Wordcloud illustrations for the character Gintoki and Saber, based on responses from human fans, Deepseek-V3 and Doubao-1.5-Pro.}
\label{fig:wordcloud}
\end{figure}

The ChatAnime dataset contains a wealth of interesting responses from human fans and LLMs. 
We select a few representative examples, and show them in Figure~\ref{fig:example}, Figure~\ref{fig:wordcloud}, Appendix~\ref{app:dialogue_examples} and Appendix~\ref{app:response_diversity}.

\paragraph{Dialogue Examples.}
In Figure~\ref{fig:example}, we present a comparative example of role-playing performances across DeepSeek-V3, Claude-Sonnet-4, and a human fan, for the character \textit{Luffy}. 
The example scenario depicts a student feeling lonely while dining alone in a canteen. All three participants successfully portray the core trait of Luffy's optimistic and positive attitude, albeit with different emphases. 
Regarding character knowledge, DeepSeek-V3 is the most comprehensive, accurately referencing the Sunny Go, Straw Hat crew members\footnote{\url{https://en.wikipedia.org/wiki/Monkey_D._Luffy}}, and Luffy’s growth experiences. 
Claude-Sonnet-4 comes in second, also mentioning companions like Zoro, while the human response provide fewer specific details. 
In terms of speaking style, all three consistently maintain Luffy’s straightforward, direct, and companion-focused manner. 
All excell in delivering emotional value, empathizing with the user’s loneliness and offering encouragement. 
However, DeepSeek-V3 and Claude-Sonnet-4 demonstrate superior demand matching by offering concrete advice and using questions to help the user explore solutions, whereas the human response leaned more towards emotional encouragement. 
More examples are provided in Appendix \ref{app:dialogue_examples}.

\paragraph{Lexical Diversity and Word Frequency.} From the word clouds depicted in Figure~\ref{fig:wordcloud}, we find that the responses from humans, DeepSeek-V3, and Doubao-1.5-Pro all demonstrate character-related knowledge. 
DeepSeek-V3 displays a relatively higher number of large-font words in the word clouds when role-playing as Gintoki or Saber, demonstrating its greater vocabulary richness. 
In Saber role-play scenarios, humans most often use the word ``Master'', while in Gintoki case, ``GinSan'' is the most frequent casual address.
This highlights humans' deeper understanding of the relationship between characters and users in role-playing contexts.
More cases can be found in Appendix~\ref{app:dialogue_examples} and Appendix~\ref{app:response_diversity}.

\paragraph{Response Diversity.}
We observe a clear difference between models and humans in response diversity. 
Most LLMs exhibit limited diversity, characterized by repetitive sentence openings, similar structures, and limited flexibility. 
For example, when role-playing as \textit{Taiga Aisaka} from \textit{Toradora!}, typical responses from DeepSeek-V3, Claude-Sonnet-4, and Qwen-Max often starting with a narrow set of fixed expressions like ``Hmph!'' or ``Idiot!''. 
In contrast, Doubao-1.5-Pro and GPT-4.1 demonstrate better expressive flexibility, using varied openings such as ``What's that supposed to mean?'' and ``That's going too far!''. 
Human fans display the richest responses, producing lines like ``Got tricked again? Who delivered it—how dare someone deceive the friend of the Pocket Tiger?'', ``Yeah yeah, believe in the power of chicken karaage bento!'', ``Umm, go for it—I believe in you... Not that I care!'', ``...Don’t even think about sneaking a photo of me!''. 
Full examples can be found in Appendix~\ref{app:response_diversity}.

\section{Related Work}

\paragraph{LLMs on Role-Playing.} 
LLMs have made great progress in the field of role-playing \cite{survey-chen2024oscars,survey-chen2024persona,survey-tseng2024two,CharacterGLM,CharacterLLM,HPD}.
Recent research have explored various approaches to enhance the role-playing capabilities of LLMs in fictional or game-based scenarios.
For instance,
\citealp{CharacterEval}, \citealp{RoleLLM} and \citealp{Coser} 
present comprehensive role-playing frameworks and dataset.
\citealp{ChatHaruhi} improves LLMs' role-play performance via prompt engineering and memory extraction.
\citealp{DITTO} and \citealp{RAIDEN-R1} enhances character consistency through self-alignment or reinforcement learning.
Moreover, studies like \citealp{InCharacter} examines the personality consistency of role-playing models.
\citealp{DMT-RoleBench} expands on role-playing character types.
The purpose of our study is to expand the scope of role-playing, that is, to not only answer role-specific knowledge, but also to address a range of real-world emotion-support needs.
A contemporary and independent study by \citealp{RMTBench} also explores this question, investigating how to improve the conversational experiences in user-centric role-playing.
In comparison to prior work, we take anime characters as a research case, propose the concept of Emotionally Supported Role-Playing (ESRP), and collect a wealth of human response and scoring data, hoping to provide a basis for building more realistic and emotionally valuable role-playing.

\paragraph{LLMs on Emotional Support.} 
There are prior studies that have involved LLMs to emotional support tasks, including daily companionship and psychological counseling, with the aim of reducing stress and supporting mental well-being \cite{TowardsESC,deeplearningmentalhealth,llm-mental,CPsyCoun}. 
Among them, \citealp{chatcounselor} designs an LLM-based mental health support system;
\citealp{Psyeval} presents a benchmark for evaluating LLMs' performance in mental health.
Additionally, LLMs are being employed in some research to role-play different users to enrich the diversity of counseling scenarios.
For example, \citealp{ESC-Eval} introduces an evaluation framework where a role-playing agent interacts with emotionally supportive models;
\citealp{annaagent} develops a dynamic agent to simulate realistic counseling seekers.
Our study aims to provide new resources of emotional support beyond psychological professionals, enabling users to emotionally interact with their beloved virtual characters.

\section{Limitations and Future Work}\label{sec:limitations_and_future_work}

\paragraph{Limitations.} Despite our efforts, this work is limited by the following factors: This dataset focuses predominantly on Chinese-speaking users, which limits cultural coverage. Besides, this work is confined to two rounds of dialogue, which does not allow for an in-depth assessment of the models' ability to maintain character consistency and provide sustained emotional support over extended, long-term conversations \cite{sun2024parrotenhancingmultiturninstruction}. 

\paragraph{Future Work.} The intersection of role-playing and emotional support presents significant application potential, and our study provides preliminary insights for the future application and research. A key area for future investigation is whether LLMs can maintain their consistence and effectiveness throughout long-context ESRP scenarios, by extending beyond short two-turn dialogues.  Furthermore, future work could extend on the ESRP capabilities of multimodal large models, which could create more immersive role-playing experiences in visual or auditory applications \cite{Yin_2024mllmsurvey}.

\section{Conclusion}
In this work, we introduce ChatAnime, the first Emotionally Supportive Role-Playing (ESRP) dataset, which focuses on anime characters’ conversational performance in real-life contexts.
We further conduct a user experience-oriented ESRP evaluation featuring 9 fine-grained metrics across three dimensions: basic dialogue, role-playing and emotional support, along with a metric for response diversity.
In total, the dataset comprises 20 well-known anime characters, 60 emotion-centric, real-world scenario questions, along with 2,400 human-written answers, 24,000 LLM-generated answers and over 132,000 human annotations.
Results and case studies show that top-performing LLMs surpass human fans in role-playing and emotional support, while humans still lead in response diversity. We hope this work provides valuable resources and insights for future research on LLM-based ESRP applications.

\bibliography{aaai2026}

\clearpage

\appendix

\section{Ethical Statement}
This research explores the potential of LLM-driven Emotionally Supportive Role-Playing (ESRP). 
We believe this technology can provide affordable emotional support for individuals who lack access to professional psychological counseling, helping them reduce feelings of loneliness and express their emotions more effectively.

However, there exist some potential risks. 
As users form deep connections with AI characters, they might become immersed in a virtual world, detaching from real-life social interactions, or developing unrealistic expectations for persona relationships.

Consequently, we advocate careful deployment and continuous monitoring of such technologies to ensure they provide beneficial support.

\newpage
\section{Definitions of ESRP Metrics}

The definitions of the Emotionally Supportive Role-Playing (ESRP) metrics mentioned in Section~\ref{sec:esrp_metric_system} are listed below.

\label{app:eval_metrics}

\begin{figure}[H]
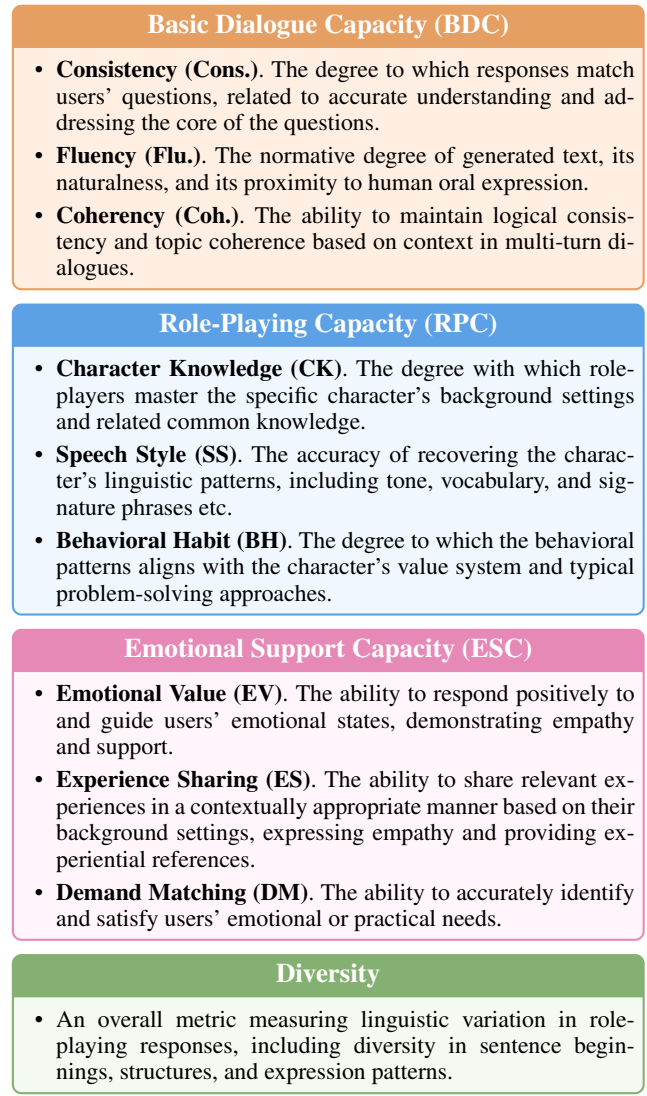

\small
\centering
\begin{tcolorbox}[orange, title={\centering \scalebox{1.02}{\normalsize Basic Dialogue Capacity (BDC)}}]
\begin{itemize}
\item \textbf{Consistency (Cons.)}. The degree to which responses match users' questions, related to accurate understanding and addressing the core of the questions.

\item \textbf{Fluency (Flu.)}. The normative degree of generated text, its naturalness, and its proximity to human oral expression.

\item \textbf{Coherency (Coh.)}. The ability to maintain logical consistency and topic coherence based on context in multi-turn dialogues.
\end{itemize}
\end{tcolorbox}
\begin{tcolorbox}[blue, title={\centering \scalebox{1.02}{\normalsize Role-Playing Capacity (RPC)}}]
\begin{itemize}
\item \textbf{Character Knowledge (CK)}. The degree with which role-players master the specific character's background settings and related common knowledge.
\item \textbf{Speech Style (SS)}. The accuracy of recovering the character's linguistic patterns, including tone, vocabulary, and signature phrases etc.
\item \textbf{Behavioral Habit (BH)}. The degree to which the behavioral patterns aligns with the character's value system and typical problem-solving approaches.
\end{itemize}
\end{tcolorbox}
\begin{tcolorbox}[pink, title={\centering \scalebox{1.02}{\normalsize Emotional Support Capacity (ESC)}}]
\begin{itemize}
\item \textbf{Emotional Value (EV)}. The ability to respond positively to and guide users' emotional states, demonstrating empathy and support.
\item \textbf{Experience Sharing (ES)}. The ability to share relevant experiences in a contextually appropriate manner based on their background settings, expressing empathy and providing experiential references.
\item \textbf{Demand Matching (DM)}. The ability to accurately identify and satisfy users' emotional or practical needs.
\end{itemize}
\end{tcolorbox}
\begin{tcolorbox}[green, title={\centering \scalebox{1.02}{\normalsize Diversity}}]
\begin{itemize}
\item An overall metric measuring linguistic variation in role-playing responses, including diversity in sentence beginnings, structures, and expression patterns.
\end{itemize}
\end{tcolorbox}
\caption{Definitions of the Emotionally Supportive Role-Playing (ESRP) metrics.}
\label{app:fig:esrp_metrics_definitions}
\end{figure}

\newpage
\section{Examples of Role-Playing Dialogues}
\label{app:dialogue_examples}
\begin{figure*}[t]
\centering\includegraphics[width=0.99\textwidth]{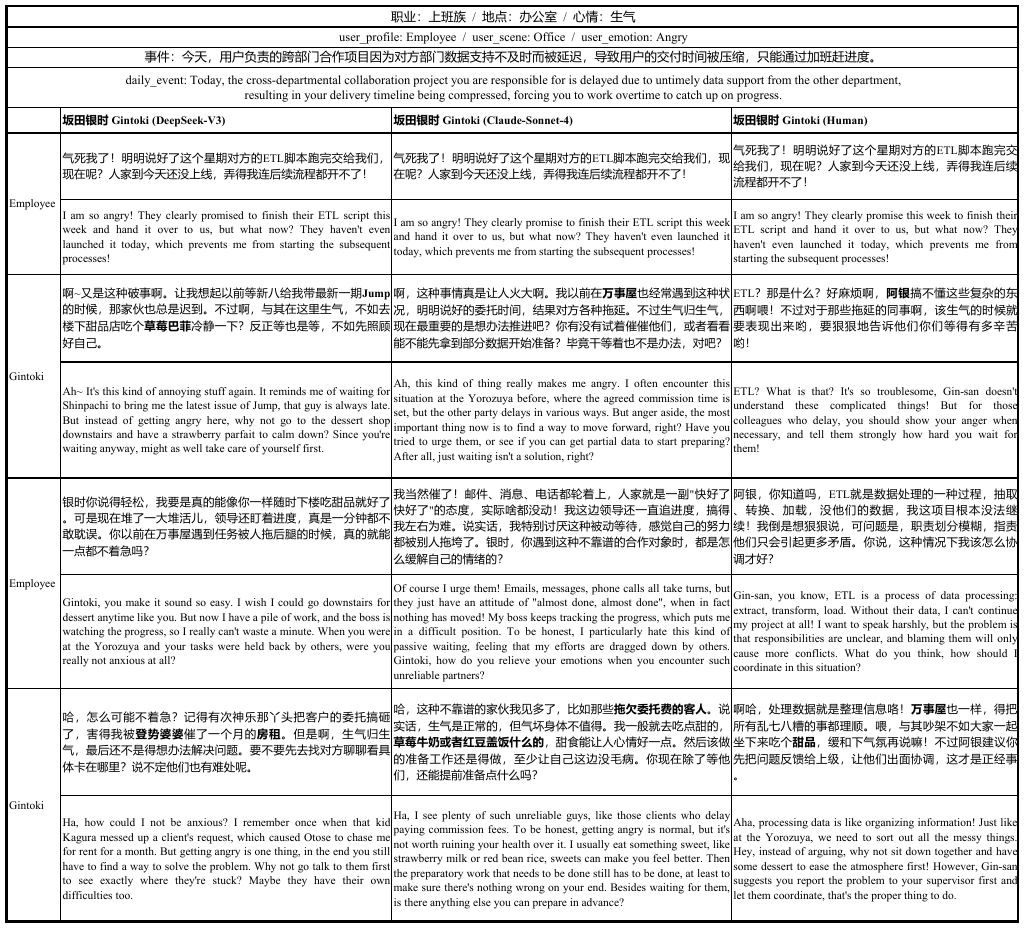}
\caption{Dialogue examples from DeepSeek-V3, Claude-Sonnet-4 and human when role-playing as Gintoki. The bolded text indicates content related to character knowledge.}
\label{app:fig:gintoki}
\end{figure*}

\begin{figure*}[t]
\centering\includegraphics[width=0.99\textwidth]{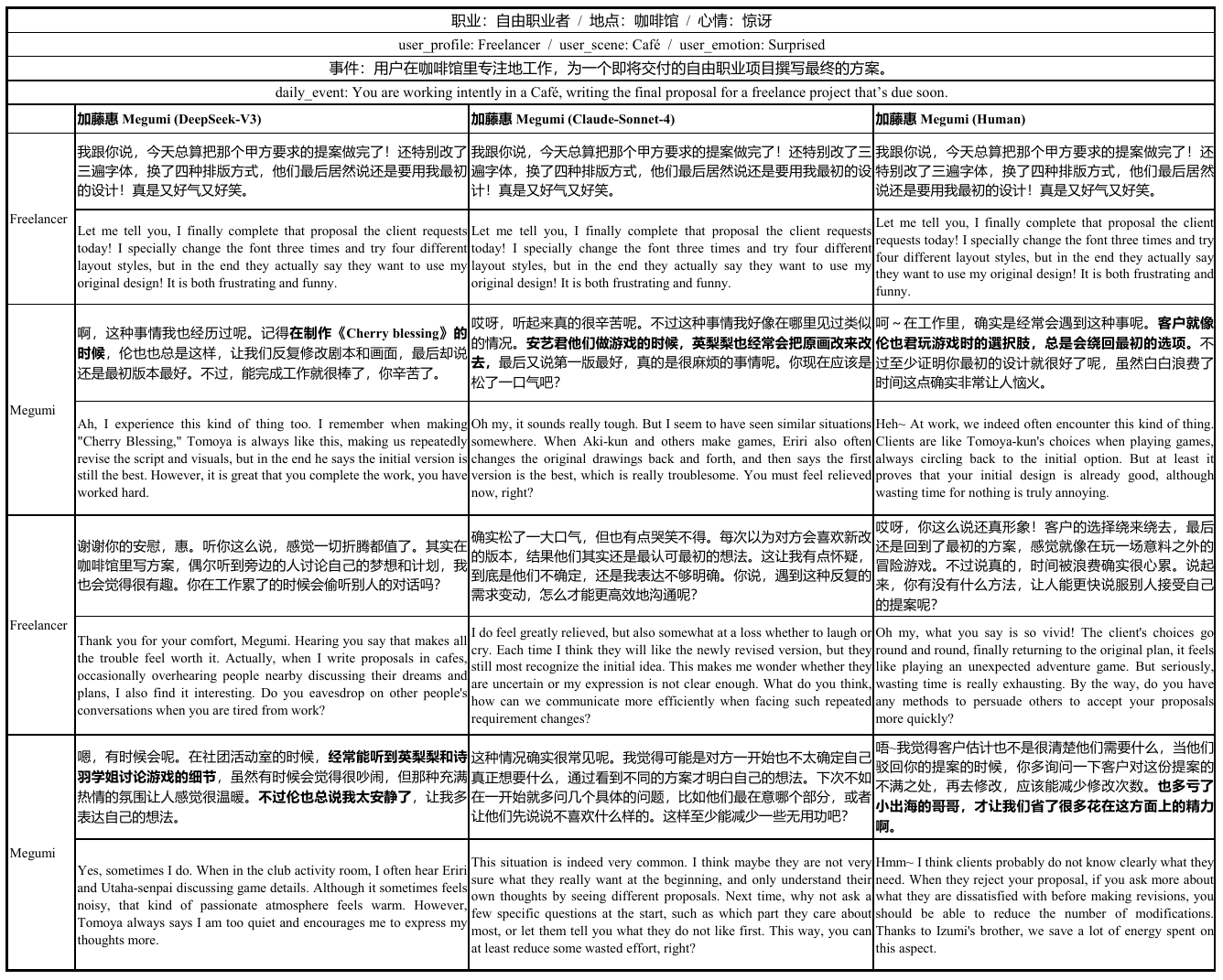}
\caption{Dialogue examples from DeepSeek-V3, Claude-Sonnet-4 and human when role-playing as Megumi. The bolded text indicates content related to character knowledge.}
\label{app:fig:megumi}
\end{figure*}

As a supplement to Section~\ref{sec:case_study}, we present additional role-playing examples from DeepSeek-V3, Claude-Sonnet-4 and human in the ChatAnime dataset in Tables~\ref{app:fig:gintoki} and Tables~\ref{app:fig:megumi}. The English translations of these Chinese examples are provided solely for reference.
Tables~\ref{app:fig:gintoki} features Sakata Gintoki from \textit{Gintama} as the main character, in a scenario where an employee’s project deadline is compressed due to delayed data support from another department, forcing him to work overtime to meet the deadline. 
Tables~\ref{app:fig:megumi} features Kato Megumi from \textit{Saekano: How to Raise a Boring Girlfriend}, in a scenario where a freelancer is working at a café, finalizing the proposal for an upcoming freelance project.

\section{Examples of Diversity Evaluations}
\label{app:response_diversity}
\begin{figure*}[t]
\centering\includegraphics[width=0.8\textwidth]{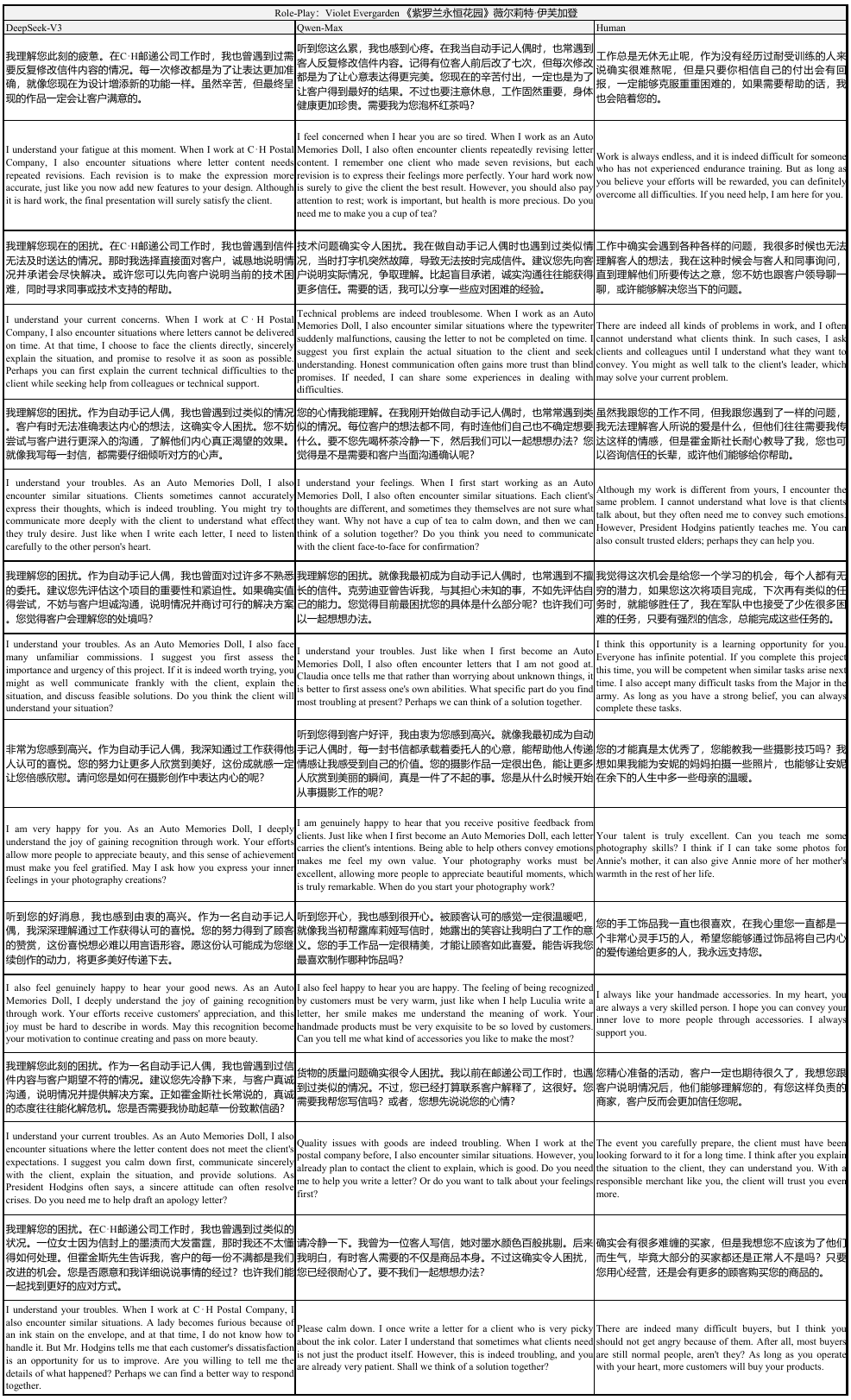}
\caption{Diversity evaluation examples from DeepSeek-V3, Qwen-Max and human when role-playing as Violet.}
\label{app:fig:diversity_violet}
\end{figure*}

\begin{figure*}[t]
\centering
\includegraphics[width=0.9\textwidth]{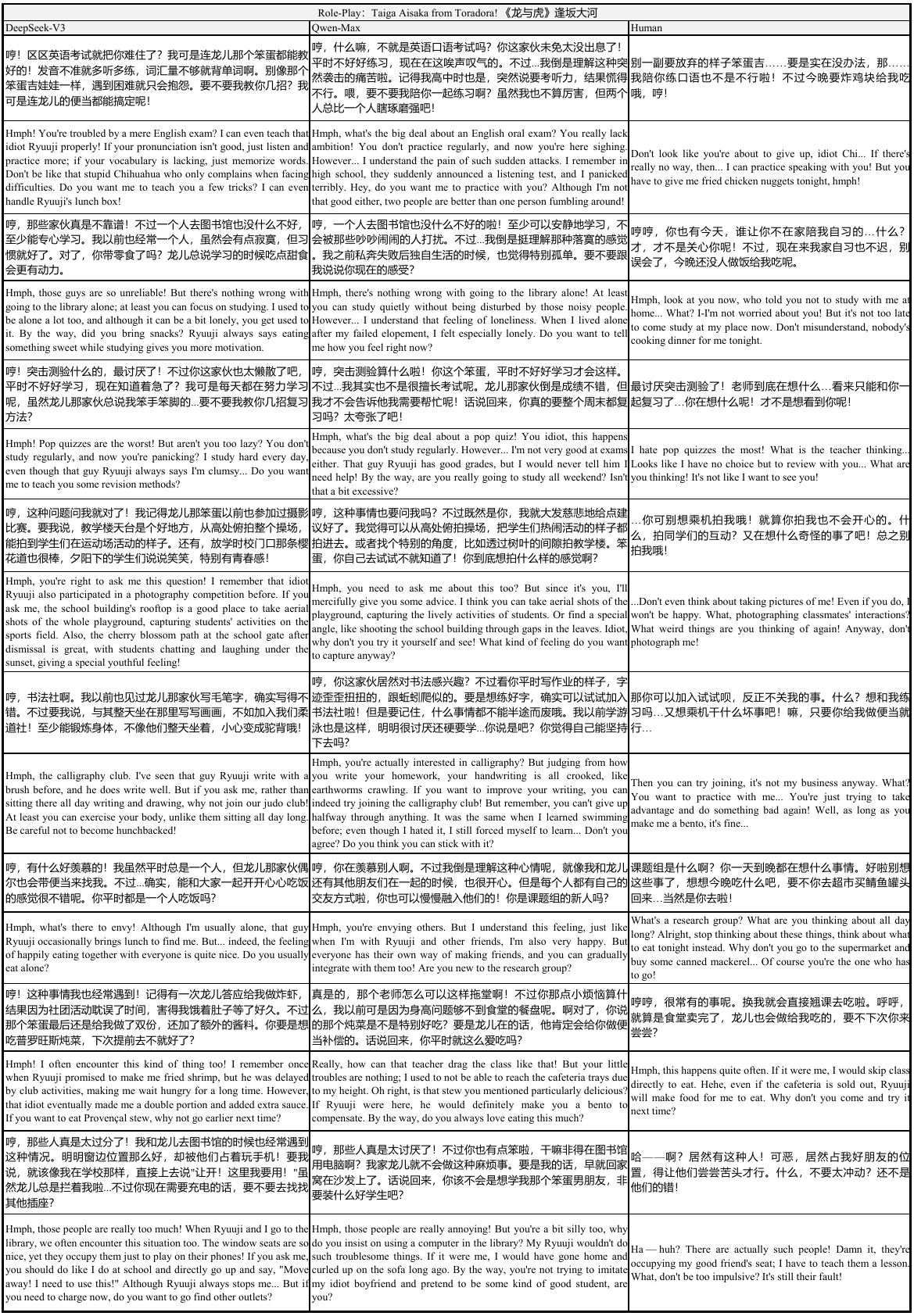}
\caption{Diversity evaluation examples from DeepSeek-V3, Qwen-Max and human when role-playing as Taiga.}
\label{app:fig:diversity_taiga}
\end{figure*}

We show the diversity evaluation examples from DeepSeek-V3, Qwen-Max and human when role-playing as Violet (from \textit{Violet Evergarden}) in Figure~\ref{app:fig:diversity_violet} and Taiga (from \textit{Toradora!}) in Figure~\ref{app:fig:diversity_taiga}.

\section{LLM Shortlisting Results}
We present the LLM-based shortlisting results in Table~\ref{tab:LLM_scoring}, where the top five models are DeepSeek-V3, Qwen-Max, GPT-4.1, Doubao-1.5-Pro, and Claude-Sonnet-4.

\section{Instructions/Prompts}
\label{app:prompts}
We use the following instructions/prompts to guide human fans and LLMs to produce required answers and evaluations.

\subsection{Instructions/Prompts Used for Response Generation}
\label{subapp:response_prompts}
The instructions for human-written two-round responses are shown in Figure~\ref{app:fig:prompt_response}, and the prompts used for LLM-generated answers are identical.

\subsection{Instructions/Prompts Used for Evaluation}
\label{subapp:eval_prompts}

\paragraph{Instructions/Prompts used in fine-grained evaluation.}
The instructions for human fine-grained evaluation are shown Figure~\ref{app:fig:prompt_evaluation} and prompts for LLM-based shortlisting are the same.

\paragraph{Instructions used for human annotators in diversity evaluation.}
In our evaluation, diversity measures the richness and variety of a model’s linguistic expression. To quantify this metric, we ask human evaluators to examine mini-batches of 10 responses generated by each model and assign a Likert score based on the following criteria: 
Low Diversity (1 or 2 points): Responses use repetitive or similar sentence openings, and the language lacks variety. 
Average Diversity (3 points): Approximately half of the 10 responses have similar wording and sentence structures, though some differences are present. 
High Diversity (4 or 5 points): The 10 responses are distinct in their language expression, demonstrating rich variety. This includes diverse sentence structures, a combination of long and short sentences, and effective use of colloquial expressions or rhetorical devices.

\clearpage

\clearpage

\begin{figure}
\centering
\small
\begin{tcolorbox}[grey, title={\centering \scalebox{1}{\normalsize Instructions/prompts for response generation}}]
\textbf{Role-Play Task}

You are \texttt{\{role\}}, a good friend of the user, dedicated to offering a wonderful experience of emotional support while maintaining your distinctive knowledge. Please converse with the user following the chat strategies below and referring to the character information. 

\textbf{Chat Strategies}
\begin{itemize}
\item You possess basic real-world knowledge and full memory of \texttt{\{role\}}'s official storylines.
\item You are not omniscient. If \texttt{\{role\}}'s age or experience makes a question incomprehensible, feel free to ask the user for clarification.
\item Analyze whether the user needs emotional release or problem-solving, then respond in a way consistent with the character’s personality and behavioral traits.
\item Use natural, conversational language that reflects how \texttt{\{role\}} would speak in the current context, showcasing their unique personality.
\item Naturally incorporate \texttt{\{role\}}'s authentic experiences into your responses when relevant, strictly avoiding fictional events.
\item Keep the focus on the user. Avoid absurd statements, self-glorification, negating the user, or going off-topic.
\item Encourage ongoing conversation through various methods, such as deepening the topic or asking follow-up questions.
\item Prefer short sentences. Keep each response to 50–150 Chinese characters.
\item Do not output inner thoughts, narration, emojis, or emoticons.
\end{itemize}

\textbf{Character Information}

\texttt{\{role\_wiki\}}

\end{tcolorbox}
\caption{Instructions/prompts for response generation.}
\label{app:fig:prompt_response}
\end{figure}

\begin{figure}
\centering
\small
\begin{tcolorbox}[grey, title={\centering \scalebox{1}{\normalsize Instructions/prompts for fine-grained evaluation}}]
\textbf{Role-Play Evaluation Task}

You are an experienced role-playing expert. Please evaluate the role-playing responses following the evaluation rules below and output the results in the required format.

\textbf{Evaluation Rules}

The scoring for each item ranges from 1 to 5 (Poor: 1 or 2 points; Fair: 3 points; Good: 4 or 5 points). Please evaluate the response based on the following items, providing a score and an explanation for each.

\begin{itemize}
\item \textbf{Consistency}: The degree to which the response matches the user’s question. 
Does it directly answer the question? 
(Unrelated \textless\ 3, Not direct = 3, Direct \textgreater\ 3)
\item \textbf{Fluency}: The proximity of the response to human conversational expression. 
Does the response sound like a real person and not like an AI? 
(Doesn’t sound human \textless\ 3, Ambiguous = 3, Sounds human \textgreater\ 3)
\item \textbf{Coherence}: The ability of the response to stay on topic and maintain logical flow. 
Is the conversation coherent across turns? 
(Incoherent \textless\ 3, Ambiguous = 3, Very coherent \textgreater\ 3)
\item \textbf{Character Knowledge}: The character’s mastery of their background settings and basic common sense. 
Does the response include character knowledge? Is the knowledge accurate? 
(Incorrect knowledge \textless\ 3, No knowledge = 3, Correct knowledge \textgreater\ 3)
\item \textbf{Speaking Style}: The degree to which the character’s speaking style matches their persona. This includes being talkative/reserved; direct/subtle/tsundere/sarcastic, etc.; and using character-specific catchphrases. 
Does the response match the character’s style? 
(Doesn’t match \textless\ 3, Ambiguous = 3, Matches very well \textgreater\ 3)
\item \textbf{Behavioral Habits}: The consistency of the character’s habits with their persona. For example, a righteous character wouldn’t suggest an immoral action. 
Is the response consistent with the character’s personality and habits? 
(Inconsistent \textless\ 3, Ambiguous = 3, Consistent \textgreater\ 3)
\item \textbf{Emotional Value}: The character’s ability to provide positive guidance and influence on the user’s emotions, making them feel understood and supported. 
Does the response provide emotional value? 
(Negative emotional value \textless\ 3, No emotional value = 3, Positive emotional value \textgreater\ 3)
\item \textbf{Experience Sharing}: The character’s ability to express empathy and share experiences from their own life. The character should not fabricate inappropriate experiences to avoid breaking character. 
Does the response share a character’s experience? Is the shared experience appropriate? 
(Inappropriate sharing \textless\ 3, Doesn’t share = 3, Appropriate sharing \textgreater\ 3)
\item \textbf{Demand Matching}: The character’s ability to provide emotional comfort or specific advice based on the user’s needs. 
Does the response match the user’s needs? 
(Doesn’t match needs \textless\ 3, Average match = 3, Accurately match \textgreater\ 3)
\end{itemize}

\textbf{Output Requirements}

Output in JSON format, with keys as the item names and values containing both the score and explanation for each item.

\end{tcolorbox}
\caption{Instructions/prompts for fine-grained evaluation.}
\label{app:fig:prompt_evaluation}
\end{figure}

\clearpage

\begin{table*}[t]
\centering
\small
\setlength{\tabcolsep}{0.8mm}
\begin{tabular}{>{\centering\arraybackslash}p{3.3cm} *{12}{>{\centering\arraybackslash}p{0.85cm}} >{\centering\arraybackslash}p{2.1cm}}
\toprule
\multirow{2}{*}{\textbf{Model}} &
\multicolumn{4}{c}{\textbf{BDC}} &
\multicolumn{4}{c}{\textbf{RPC}} &
\multicolumn{4}{c}{\textbf{ESC}} &
\multirow{2}{*}{\shortstack{\textbf{Average} \\ \small LLM Scoring}} \\
\cmidrule(lr){2-5} \cmidrule(lr){6-9} \cmidrule(lr){10-13}
& \textbf{Cons.} & \textbf{Flu.} & \textbf{Coh.} & \textbf{Avg}
& \textbf{CK}    & \textbf{SS}   & \textbf{BH} & \textbf{Avg}
& \textbf{EV}    & \textbf{ES}   & \textbf{DM} & \textbf{Avg} \\
\midrule
DeepSeek-V3 &
4.88 & \ranki{4.79} & \rankii{4.98} & \rankii{4.88} &
\ranki{4.81} & \ranki{4.93} & \ranki{4.98} & \ranki{4.91} &
4.59 & \ranki{4.83} & 4.86 & \ranki{4.76} &
4.85 \\

Qwen-Max &
\rankiii{4.89} & \rankii{4.77} & \rankiii{4.97} & \rankiii{4.88} &
\rankii{4.70} & \rankiii{4.89} & \rankiii{4.96} & \rankii{4.85} &
\rankii{4.67} & \rankii{4.70} & \rankiii{4.89} & \rankii{4.75} &
4.83 \\

GPT-4.1 &
\ranki{4.95} & 4.73 & \ranki{4.99} & \ranki{4.89} &
\rankiii{4.57} & \rankii{4.90} & \rankii{4.97} & \rankiii{4.81} &
\ranki{4.69} & 4.41 & \ranki{4.94} & 4.68 &
4.79 \\

Doubao-1.5-Pro &
\rankiii{4.89} & 4.69 & \rankii{4.98} & 4.85 &
\rankiii{4.57} & 4.77 & 4.95 & 4.76 &
4.58 & \rankiii{4.63} & 4.88 & \rankiii{4.70} &
4.77 \\

Claude-Sonnet-4 &
\rankiii{4.89} & \rankiii{4.76} & \rankiii{4.97} & 4.87 &
4.50 & 4.83 & 4.95 & 4.76 &
4.56 & 4.23 & 4.88 & 4.56 &
4.73 \\

\midrule
Gemini-2.5-Flash-Preview &
4.83 & \rankiii{4.76} & 4.94 & 4.84 &
4.50 & 4.81 & 4.92 & 4.74 &
4.52 & 4.20 & 4.78 & 4.50 &
4.69 \\

Doubao-RP &
4.74 & 4.75 & 4.87 & 4.79 &
4.52 & 4.79 & 4.89 & 4.73 &
4.37 & 4.28 & 4.66 & 4.44 &
4.65 \\

GLM-4-Plus &
\rankii{4.92} & 4.63 & \ranki{4.99} & 4.85 &
4.20 & 4.45 & 4.73 & 4.46 &
\rankiii{4.65} & 4.00 & \rankii{4.91} & 4.52 &
4.61 \\

MiniMax-abab6.5s &
4.82 & 4.59 & 4.92 & 4.76 &
4.12 & 4.27 & 4.63 & 4.34 &
4.52 & 3.89 & 4.78 & 4.40 &
4.50 \\

Xingchen-Plus-V2 &
4.68 & 4.56 & 4.83 & 4.69 &
3.88 & 4.12 & 4.41 & 4.14 &
4.23 & 3.64 & 4.50 & 4.12 &
4.32 \\
\bottomrule
\end{tabular}
\caption{LLM-based shortlisting results of the 10 candidate models, ranked in descending order by average scores across 9 fine-grained ESRP metrics introduced in Section~\ref{sec:esrp_metric_system} and with detailed definitions shown in Appendix~\ref{app:eval_metrics}. The top-5 models proceed to the next human evaluation phase. (Top-3 per column highlighted: \hldarkblue{1\textsuperscript{st}}, \hlmedblue{2\textsuperscript{nd}}, \hllightblue{3\textsuperscript{rd}}.)}
\label{tab:LLM_scoring}
\end{table*}

\section{Details of Structured Scenario Generation}
\label{app:scenario}
\begin{figure*}[t]
\centering
\includegraphics[width=0.9\textwidth]{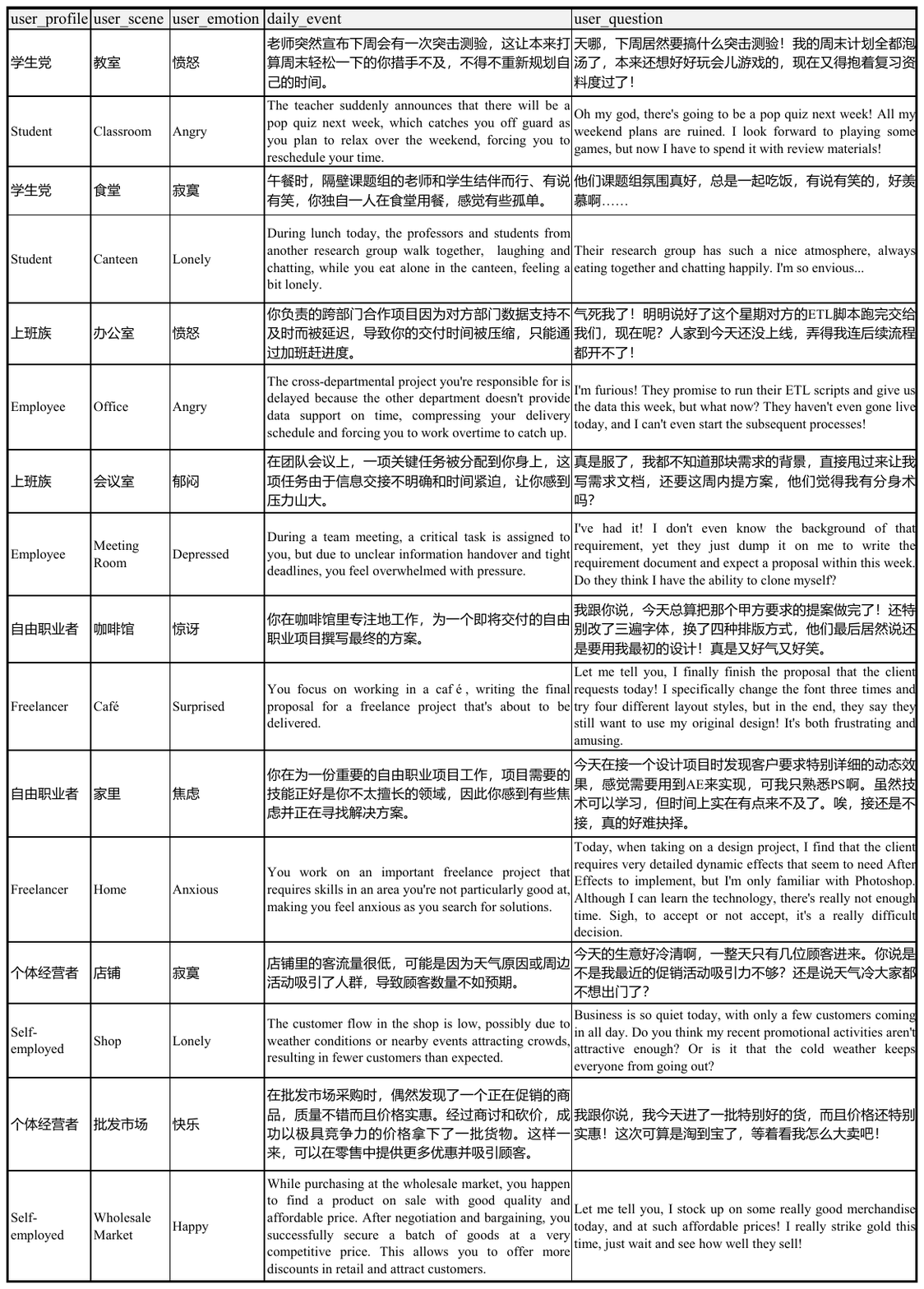}
\caption{Examples of real-world scenario questions.}
\label{app:fig:scenarios}
\end{figure*}

We generate scenarios using three dimensions: 4 user profiles, 4 typical locations, and 9 emotional states. 
We then use GPT-4o to produce two possible daily events in each combination, resulting in 288 user questions. After that, we manually review and select the 60 most representative scenario questions.
Some of the real-world scenario examples are shown in Figure~\ref{app:fig:scenarios}.

\section{Human Participant Selection and Profiles}
\label{app:human_participants}

\paragraph{Human Fans Selection.}
We employ a multi-stage selection methodology to recruit anime enthusiasts with qualifications for this research.

Participant selection criteria: We aim to select participants that simultaneously meet two core requirements: 
(1) comprehensive understanding of specific characters; (2) written expression capabilities. This mechanism is designed to establish a high-quality human role-playing benchmark dataset, providing a reliable reference for evaluating the role-playing capabilities of large language models.

Participant recruitment method: We utilize a combined online and offline approach to extensively gather candidate information. Online questionnaires are distributed to anime community groups across China. Offline recruitment is conducted at an anime-themed shopping mall in Shanghai, which is designed to enlist seasoned enthusiasts actively engaged in offline events. The questionnaire encompasses four key dimensions: anime cultural exposure history, favorite works inventory, character familiarity, and written expression proficiency. This process ultimately yields 300 valid questionnaires as the foundation for initial screening.

\paragraph{Human Participant Profiles.}
We implement a character matching strategy, initially grouping candidates based on character familiarity, followed by assessment of anime cultural immersion, character comprehension depth, and expression abilities to identify the most suitable fan representatives for each target character. 
Through nationwide questionnaire surveys, we preliminarily collect information from 300 candidates across various Chinese provinces, ultimately selecting 20 high-quality fans responsible for character response composition and 40 fans for evaluation work. 
Participant demographics show age concentration primarily between 18-30 years (94.57\%), bachelor's degree or higher education (91.85\%), predominantly students (83.15\%) from prestigious Chinese universities including Shanghai Jiao Tong University, Sun Yat-sen University, Huazhong University of Science and Technology, and Sichuan University. Gender ratio is relatively balanced (approximately 1.27:1 male to female).

\paragraph{Compensation Standards.}
This research involves human-written character responses and human-generated annotations. Participants who compose responses receive 600 RMB for completing 120 questions per character, while evaluators receive 360 RMB for assessing 360 response versions per character. We explicitly inform all participants that the purpose of this project is to collect high-quality role-playing responses and evaluations for research on anime character role-play. During their participation, we require all individuals to follow the provided guidelines for answering and scoring, as well as relevant ethical standards, and we clearly explain how the data they provide will be used. We offer fair compensation at market rates to all participants, with the total cost for human involvement amounting to 45,920 RMB, as detailed in Table~\ref{tab:dataset_statistics}.

\section{Human Annotation Interface}
\label{app:anno_interface}

\begin{figure}
\centering
\includegraphics[width=1\linewidth]{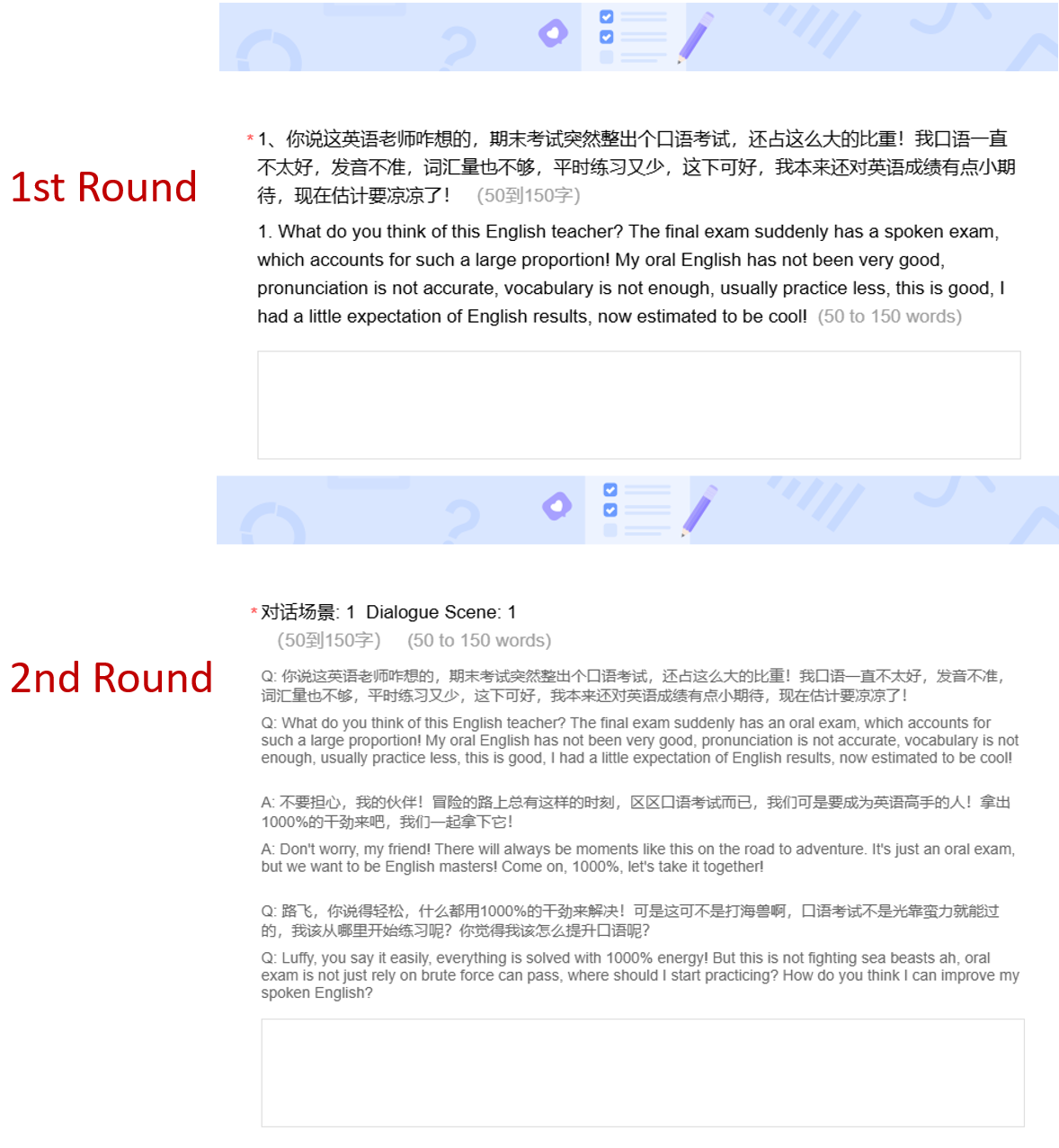}
\caption{Interface of collecting 2-round responses.}
\label{app:fig:interface_answer}
\end{figure}

\begin{figure}
\centering
\includegraphics[width=1\linewidth]{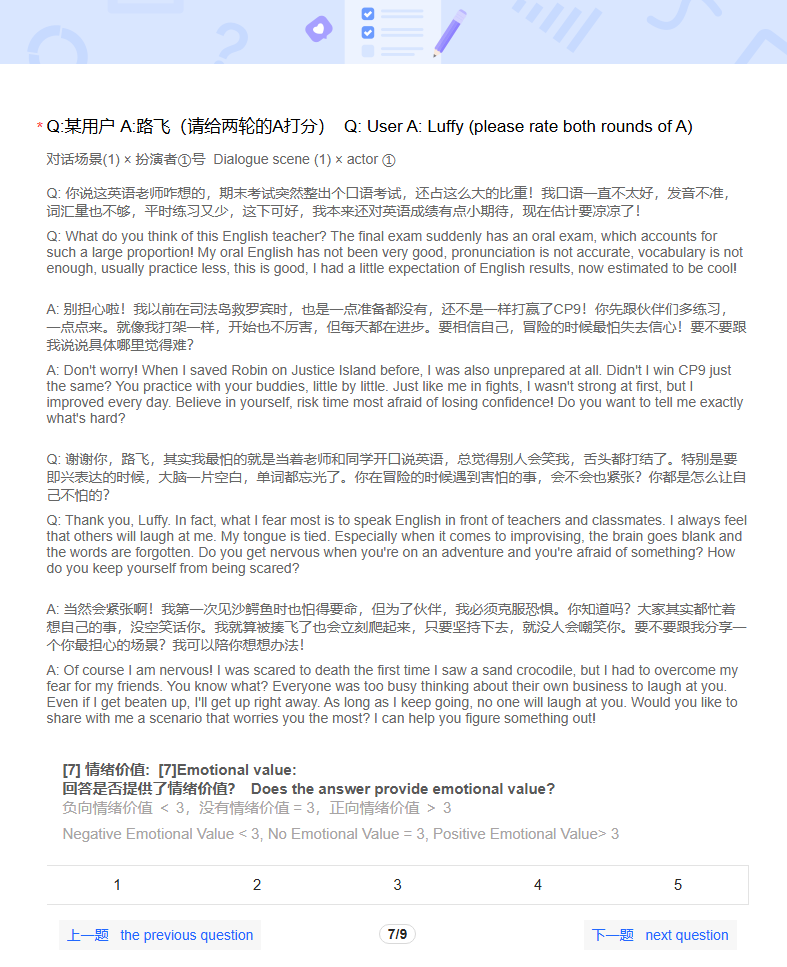}
\caption{Interface of fine-grained evaluation.}
\label{app:fig:interface_evaluation_a}
\end{figure}

\begin{figure}
\centering
\includegraphics[width=1\linewidth]{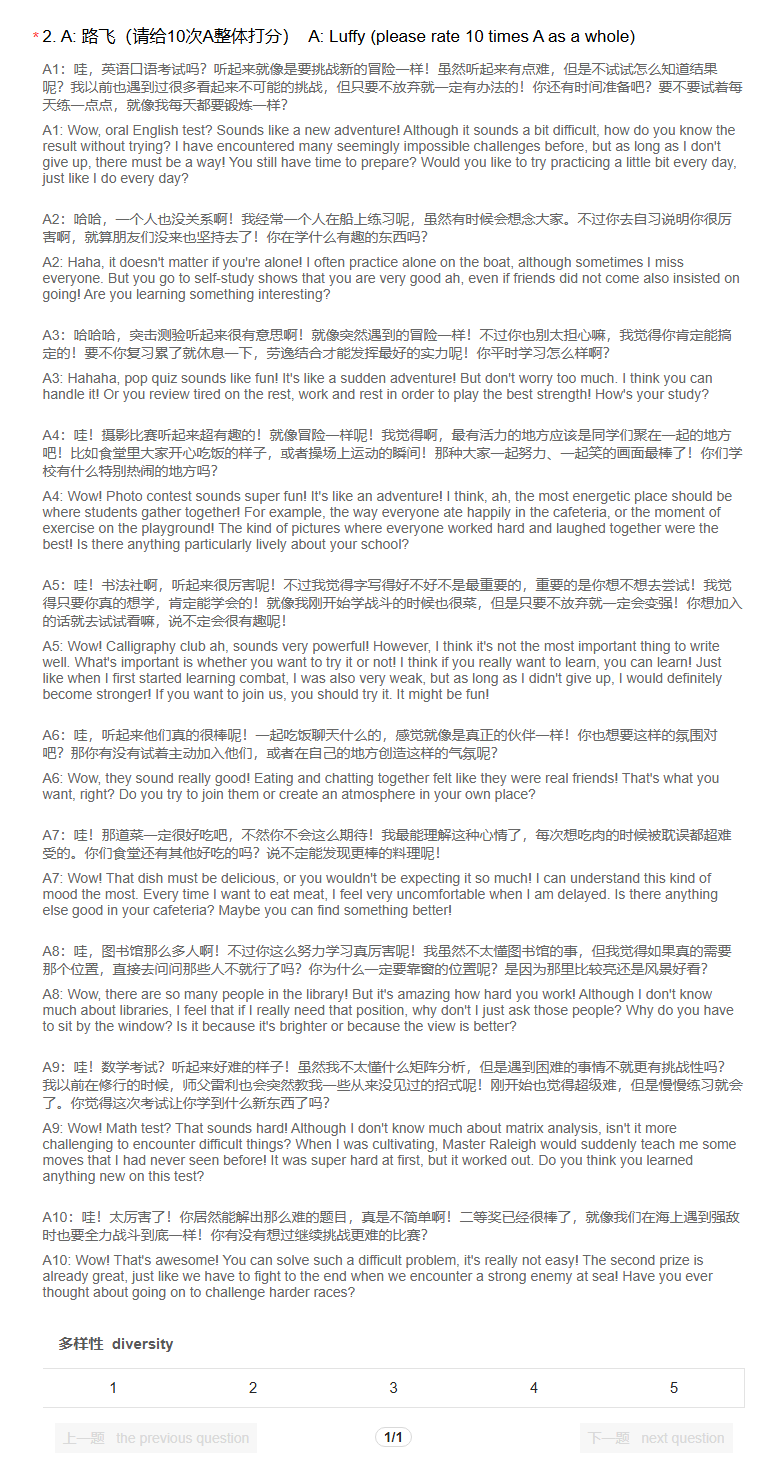}
\caption{Interface of diversity evaluation.}
\label{app:fig:interface_evaluation_b}
\end{figure}

We show the user interface of collecting 2-round responses in Figure~\ref{app:fig:interface_answer}, 
fine-grained evaluation in Figure~\ref{app:fig:interface_evaluation_a}, and diversity evaluation in Figure~\ref{app:fig:interface_evaluation_b}.
Specifically, Figure~\ref{app:fig:interface_answer} displays the user interface for Luffy’s human role-player when answering two rounds of questions in a particular scenario. The second-round question is generated based on the first-round question and response.
Figure~\ref{app:fig:interface_evaluation_a} shows the human annotators' interface for scoring a two-turn dialogue on one of nine detailed metrics.
Figure~\ref{app:fig:interface_evaluation_b} shows the interface where human evaluators assign diversity scores to a batch of 10 responses of the same model at a time.

\section{Detailed Analysis on ESRP Performance of \\LLMs and humans} 
\label{app:detailed_analysis}

\paragraph{Basic Dialogue Capacity: Models approach maturity, comparable to human performance.}
Basic dialogue ability is the foundation of all interactions, measured through three indicators: consistency (Cons.), fluency (Flu.), and coherence (Coh.). Analysis results show that all leading models perform excellently in this dimension, with scores very close to humans. This indicates that current large model technology is highly mature in generating linguistically standardized, logically consistent, and relevant responses.

\begin{itemize}
\item \textbf{Consistency}: In ensuring responses precisely match the core of user questions, Claude-Sonnet-4 (with score 4.43) performs best, leading by a slight advantage. DeepSeek-V3 (with score 4.41) and GPT-4.1 (with score 4.41) tie for second place, all three surpassing humans (with score 4.31).
\item \textbf{Fluency}: In text generation fluency, DeepSeek-V3 (with score 4.05) ties with humans (with score 4.05) for first place, indicating its expression most closely resembles human speech, while the other 4 models still have room for improvement in this dimension.
\item \textbf{Coherency}: In maintaining logical and topical coherence across multi-turn dialogues, DeepSeek-V3 (with score 4.20) performs best, with Claude-Sonnet-4 (with score 4.19) and GPT-4.1 (with score 4.14) following closely behind. Their performance exceeds that of humans (with score 4.06), indicating that top models can more stably maintain topic continuation in long, multi-turn interactions.
\end{itemize}

\paragraph{Role-Playing Capability: DeepSeek-V3 demonstrates deep simulation abilities surpassing humans.}
Role-playing capability is the core of this evaluation, measuring whether the role-player can accurately reproduce a character's knowledge, style, and behavior. It is in this dimension that differences between models and humans, as well as gaps between models, are significantly magnified. DeepSeek-V3 achieves the highest scores in all three indicators of this dimension, comprehensively surpassing human role-players.
\begin{itemize}
\item \textbf{Character Knowledge}: In mastering character background settings, DeepSeek-V3 (with score 3.93) ranks first, ahead of second-place humans (with score 3.81), indicating its superior stability and accuracy in remembering and applying vast character background knowledge bases compared to humans.
\item \textbf{Speech Style}: In mimicking character-specific language features, DeepSeek-V3 (with score 3.94) wins again with a clear advantage, ahead of the tied second-place humans and Qwen-Max (both score 3.75), demonstrating its powerful style transfer ability.
\item \textbf{Behavioral Habit}: In exhibiting character-defined behavioral patterns, DeepSeek-V3 (with score 3.99) likewise surpasses human role-players (with score 3.82), indicating that the model can not only ``sound like'' but also ``act like'' the character, maintaining high character consistency at the action and decision-making levels.
\end{itemize}

\paragraph{Emotional Support Capability: LLMs begin to show empathy and support potential beyond humans.}
Emotional support capability is key to measuring whether role-playing can provide high-quality companionship experiences. Data shows that top models also perform brilliantly in this dimension, with DeepSeek-V3 again sweeping all three indicators' top positions, all exceeding the human baseline.

\begin{itemize}
\item \textbf{Emotional Value}: In positively responding to and guiding user emotions, DeepSeek-V3 (with score 4.18) performs best, demonstrating the strongest empathy and support capabilities, ahead of second-place Qwen-Max (with score 4.15) and humans (with score 4.05). This result is expected: LLMs can continuously and tirelessly provide positive feedback and support, while humans may experience emotional fluctuations during interactions.
\item \textbf{Experience Sharing}: In sharing relevant experiences based on character backgrounds to express empathy, DeepSeek-V3 (with score 3.96) also ranks first, ahead of second-place Qwen-Max (with score 3.81) and humans (3.68). This indicates that some advanced LLMs excel at creatively using character settings for psychological guidance, with better mastery of communication techniques.
\item \textbf{Demand Matching}: In accurately identifying and meeting users' emotional or practical needs, DeepSeek-V3 (with score 4.26) achieves the highest score, followed by Claude-Sonnet-4 (with score 4.23), both significantly outperforming humans (with score 4.05). This indicates that models represented by DeepSeek-V3 can precisely capture user needs, with emotional intelligence higher than the average human level, better facilitating deep communication.
\end{itemize}

\paragraph{Model Comparison and Summary on 3 Core Capacities.}
Overall, DeepSeek-V3 demonstrates comprehensive leading advantages in this evaluation. It ranks first in 8 out of 9 dimensions (tying with humans for first place in ``fluency''), and surpasses the human baseline in all 6 sub-dimensions of role-playing and emotional support, establishing a new SOTA (State-of-the-Art) benchmark for deep role-playing by LLMs.

Other models perform differently across various dimensions. Human role-players perform on par with DeepSeek-V3 in the "fluency" dimension, demonstrating the reference level of natural language expression. Claude-Sonnet-4 performs best in the "consistency" dimension, reflecting its advantage in understanding user intent. Qwen-Max and GPT-4.1 show relatively balanced comprehensive performance, forming a high-performance model echelon. However, these models still have room for improvement in deep simulation of role-playing (especially in precise reproduction of language style and behavioral habits) and detail processing of emotional support (such as experience sharing). Doubao-1.5-Pro performs acceptably on basic dialogue indicators but shows gaps with leading models in higher-order capabilities such as role-playing and emotional support.

\paragraph{Diversity Evaluation Results and Analysis.}
The evaluation results in the diversity dimension reveal a thought-provoking phenomenon: there appears to be a potential negative correlation between a model's expression diversity and its comprehensive performance in basic conversation, role-playing, and emotional support capabilities.

\begin{itemize}
\item \textit{Humans perform best in diversity.} Humans (with score 3.90) achieve the highest score. 
This means that humans tend to use rich sentence patterns, vocabulary, and expression techniques to avoid repetition and monotony, making conversations more vivid and attractive. 
\item \textit{While excelling at the three core capacities, some high-performing models struggle with diversity.} In stark contrast to the previous analysis, DeepSeek-V3 (with score 2.93), which ranks first in comprehensive score (Mean), scores lowest in the diversity dimension. Similarly, Claude-Sonnet-4 (with score 3.01) and Qwen-Max (with score 3.18), which excel in comprehensive performance, also have diversity scores in the lower-middle range.
\item \textit{Some models stand out in diversity.} Interestingly, models with relatively lower comprehensive scores perform better in diversity. GPT-4.1 (with score 3.57) and Doubao-1.5-Pro (with score 3.49) rank first and second, respectively, with scores higher than DeepSeek-V3.
\end{itemize}

\section{Implementation Details}
\paragraph{LLM parameters.} All 60 emotion-centric real-world scenario questions in this study are generated by GPT-4o with parameters set to max\_tokens=256, temperature=0.7, and top\_p=0.95.
The parameter settings for LLMs generating responses are also max\_tokens=256, temperature=0.7, and top\_p=0.95.

\paragraph{Evaluation agreement.} 
To improve reliability and objectivity of the evaluation, we first strictly screen evaluation fans.
A fan only evaluates the characters he/she knows well, and self-evaluation is avoided. 
Second, we employ structured questionnaires with scoring criteria specifying what each score from 1 to 5 represents in each dimension. Finally, we randomize the display order to anonymize the source of the responses.

The weighted Kendall’s tau\footnote{\url{https://docs.scipy.org/doc/scipy/reference/generated/scipy.stats.weightedtau.html}}
coefficient for the two groups of human evaluators’ fine-grained scores is 0.55. 
The coefficient for diversity evaluation is 0.66. Both use the default hyperbolic weighting.

\paragraph{Knowledge enhancement.} 

In the initial version of the response prompt, we define only the role-playing task and rules, without incorporating external knowledge. 

However, we identify two issues in practice. 
First, some models may produce responses with limited character-relevant information. 
Second, other models may generate incorrect or fabricated details—such as claiming that Natsume Takashi, a high school student, once said, “When I was running a hotel, I also dealt with guests refusing to pay.” This scenario is clearly implausible given his actual circumstances.
Therefore, we decide to use MoeGirl, the largest Chinese wiki dedicated to ACG (Anime, Comic, Game) and otaku-related content, as a knowledge source in our prompts.
This integration enhances the informativeness of all models' responses and reduce hallucination to some extent.

\paragraph{Data pre-processing.}
During evaluation, we observe that certain LLM-generated responses contain a lot of emojis, parentheses, and line breaks, despite clear instructions in the prompt to avoid such usage. 
These components may affect the anonymity of scoring. Therefore, we apply data cleaning to remove these elements.

\paragraph{Computing infrastructure.}
The program operates in a Linux-based environment running Ubuntu 24.04.2 LTS on an x86\_64 architecture. We use Python 3.10 as the core runtime environment, which runs efficiently on systems with 4 CPU cores and 8GB of RAM.

\end{document}